\documentclass[10pt,twocolumn,letterpaper]{article}

\usepackage{iccv}              

\usepackage{xcolor}
\usepackage{colortbl}
\usepackage{tcolorbox}
\tcbuselibrary{theorems}
\usepackage{multirow}
\usepackage{caption}

\usepackage{amssymb}
\usepackage{multirow}
\usepackage{fontawesome}
\usepackage{caption}
\usepackage{booktabs}
\usepackage{arydshln} 
\usepackage{wrapfig,booktabs}

\usepackage{tikz}

\usepackage{booktabs, multirow, xcolor, graphicx, colortbl, fontawesome}

\definecolor{lightblue}{RGB}{173,216,230}
\definecolor{lightyellow}{RGB}{255,255,224}
\definecolor{lightgreen}{RGB}{144,238,144}

\newcommand*\circled[1]{\tikz[baseline=(char.base)]{\node[fill=blue!30,shape=circle,draw,inner sep=0.5pt] (char) {#1};}}

\newcommand{\method}[1]{\textsc{QUestion-only replay with Attention Distillation}}

\newcommand{\qstmethod}[1]{\textsc{Question-only replay with Attention Distillation}}
\newcommand{\setting}{VQACL-QR}

\newcommand{\qstmethodshort}[1]{\textsc{QUAD}}

\definecolor{darkgreen}{rgb}{0.0, 0.5, 0.0} 

\definecolor{lightblue}{RGB}{173,216,230}
\definecolor{lightyellow}{RGB}{255,255,224}
\definecolor{lightgreen}{RGB}{144,238,144}

\usepackage{amssymb}
\usepackage{pifont}

\usepackage{tikz}           
\usepackage{pgfplots}       

\definecolor{iccvblue}{rgb}{0.21,0.49,0.74}
\usepackage[pagebackref,breaklinks,colorlinks,allcolors=iccvblue]{hyperref}

\def\paperID{28} 
\def\confName{ICCV}
\def\confYear{2025}

\title{Ask and Remember: A Questions-Only Replay Strategy for Continual Visual Question Answering}

\author{
Imad Eddine Marouf$^{1}$
\and
Enzo Tartaglione$^{1}$
\and 
Stéphane Lathuilière$^{1,2}$
\and 
Joost van de Weijer$^{3}$ \\
{\tt\small $^{1}$ LTCI, Télécom-Paris, Institut Polytechnique de Paris, France}\\{\tt\small $^{2}$Inria, LJK,  Univ.~Grenoble Alpes, France, $^{3}$Universitat Autónoma de Barcelona, Spain} 
}

\begin{document}
\maketitle
\begin{abstract}
Continual Learning in Visual Question Answering (VQACL) requires models to acquire new visual-linguistic skills (plasticity) while preserving previously learned knowledge (stability). The inherent multimodality of VQACL exacerbates this challenge, as models must balance stability across visual and textual domains while adapting to novel objects and reasoning tasks. Existing methods, primarily designed for unimodal settings, often fall short in addressing this dual requirement. In this work, we present \textit{QUestion-only replay with Attention Distillation} (\qstmethodshort{}), a novel approach for VQACL that leverages only past task questions for regularization. By eliminating the need to store visual data, \qstmethodshort{} not only reduces memory overhead, but also alleviates privacy concerns. Our method introduces a \textit{Question-only Replay} mechanism that selectively reuses prior task questions to counteract overfitting to the answer space of the current task, addressing the problem \textit{ out of answer set}. Complementing this, we propose \textit{Attention Consistency Distillation} to enforce both intra-modal and inter-modal attention consistency across tasks, preserving essential visual-linguistic associations. Extensive experiments on VQAv2 and NExT-QA demonstrate that \qstmethodshort{} significantly outperforms state-of-the-art methods, achieving robust performance in continual VQA. Code is available at: \href{https://github.com/IemProg/QUAD}{https://github.com/IemProg/QUAD}. \footnote{Work done during an internship at Computer Vision Center (CVC),
Universitat Autonoma de Barcelona.}
\end{abstract}    
\section{Introduction}
\label{sec:introduction}
Continual learning (CL) allows models to lean new skills while retaining prior knowledge, effectively mitigating catastrophic forgetting (CF)~\citep{mcclelland1995there, mccloskey1989catastrophic}. This capability is essential in dynamic real-world settings, where models must continuously adapt to evolving data while preserving previously acquired knowledge~\citep{zhang2023vqacl}. Despite significant advances in CL, most research has focused on unimodal tasks (\eg image classification)~\citep{wang2022dualprompt, wang2022continual, kirkpatrick2017overcoming, NEURIPS2023_15294ba2, marouf2024weightedensemblemodelsstrong, zhou2024expandable}.
However, real-world applications often require multimodal learning that integrates complex visual and textual reasoning. Visual Question Answering (VQA) stands as a representative multimodal task, requiring models to jointly interpret visual content and natural language. For instance, answering questions like ``\textit{What color is the car on the left?}'' or ``\textit{How many trucks are visible?}'' requires robust object recognition and a nuanced grasp of the linguistic query structure~\citep{zhang2023vqacl, schwenk2022okvqa, yang2022zero, cheng2023vindlu, Wang_2023_ICCV}.

\begin{figure}
    \centering
    \includegraphics[width=\linewidth]{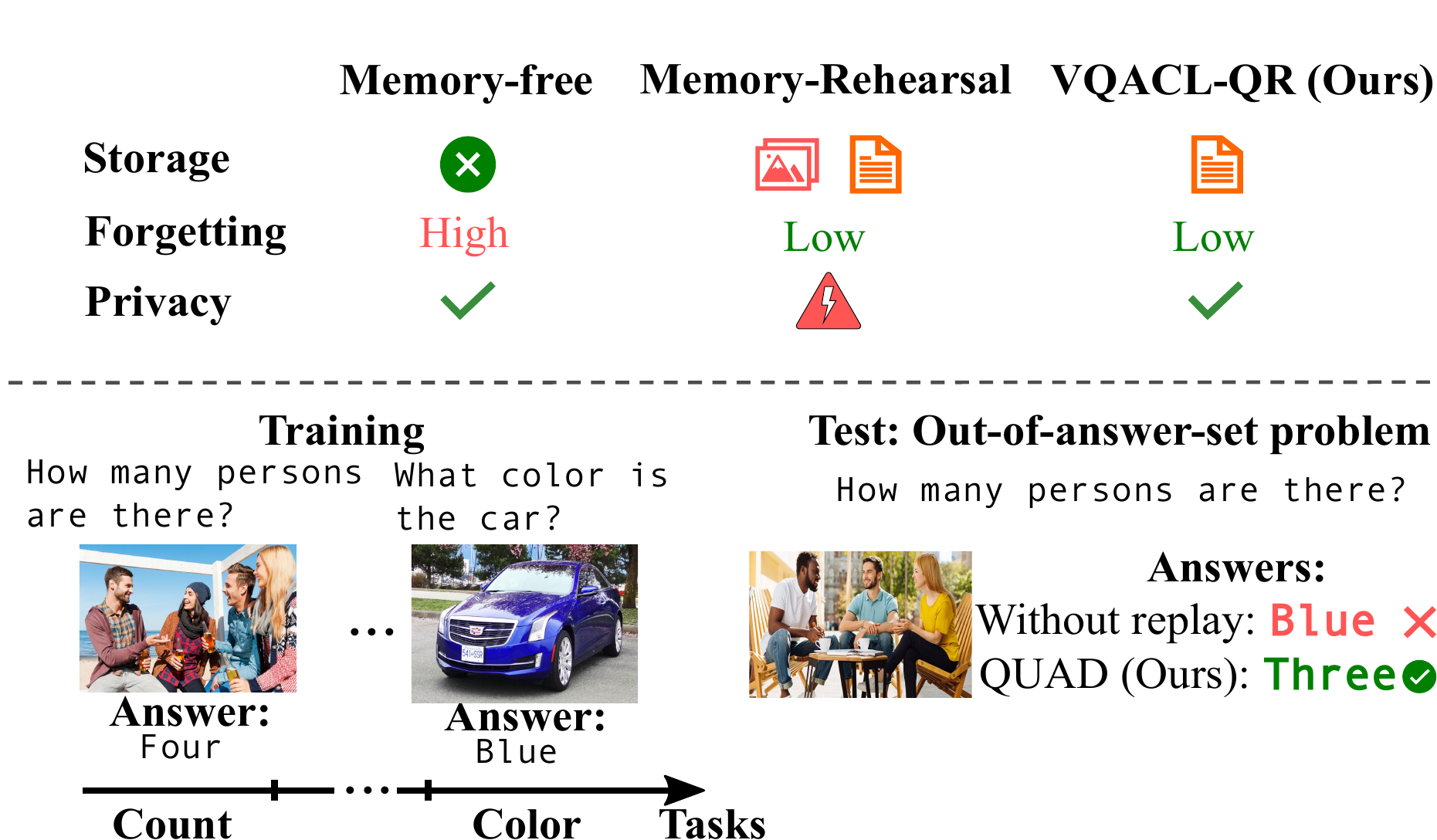}
    \caption{Comparison of continual learning methods for Visual Question Answering (VQA) in terms of storage, forgetting, and privacy. \textbf{Memory-free} methods ensure privacy but suffer from high forgetting. \textbf{Memory-rehearsal} methods reduce forgetting but raise privacy concerns by storing sensitive images (\textit{e.g.} people's identities, car plates). Our approach \textbf{\qstmethodshort{}}, only stores questions and avoids image storage, preserving privacy while achieving low forgetting by leveraging question-based regularisation to effectively solve the out-of-answer set problem.}
    \label{fig.teaser}
\end{figure}
The emerging field of Visual Question Answering Continual Learning (VQACL)~\citep{zhang2023vqacl} focuses on enabling models to improve VQA performance iteratively by learning from a sequence of tasks without catastrophic forgetting. Unlike static pretrained multimodal models~\citep{liu2023llava, li2023blip2, agrawal2024pixtral12b} that rely on massive datasets and heavy computation, VQACL offers a more efficient and scalable alternative by allowing incremental adaptation without costly re-training~\cite{zhang2023vqacl, nikandrou2022task}. Research in VQACL is further driven by the broader scientific goal of developing multimodal agents that continuously improve their reasoning throughout deployment in a multimodal environments.
A key challenge in VQACL is balancing \textit{stability}—the preservation of past knowledge—and \textit{plasticity}—the ability to learn new information—across both visual and linguistic modalities~\citep{zhang2023vqacl, antol2015vqa}. This dual-modality requirement, combined with the need for generalisation ability, introduces additional complexity in learning. Models must retain visual and linguistic knowledge across tasks, while also generalizing to novel objects and unseen question types. For example, a model that masters counting vehicles should be able to transfer this counting capability to novel object categories, such as bicycles~\citep{greco-etal-2019-psycholinguistics}.

To achieve this balance, continual VQA methods rely on memory-based replay, where previously seen examples are stored and revisited to reinforce prior knowledge and mitigate forgetting~\citep{chaudhry2019tinyepisodicmemoriescontinual, Chaudhry2019er, Wan_2022_CVPR, zhang2023vqacl}. In VQACL, this necessitates storing full image-question pairs, which demands substantial memory resources, often amounting to thousands of samples per task (\eg 5000 samples in VQACL~\citep{zhang2023vqacl}). Storing visual data poses two core challenges. \emph{First}, it incurs high computational and storage overhead, which can be infeasible in resource-constrained real-world applications. \emph{Second}, more importantly, it also raises serious privacy concerns. Visual data often contains sensitive and personally identifiable information, especially in domains like healthcare, finance, and surveillance~\citep{tian2024privacy, liu2022privacy}, where stringent regulations such as GDPR~\citep{gdpr2016general, 10.1145/3389685} govern data usage and storage. In contrast, textual data—such as questions—is typically generic and non-identifiable, thereby posing minimal privacy risks. Memory-free methods~\citep{kruengkrai-yamagishi-2022-mitigating, mas2018, Zhang_2024_CVPR, kirkpatrick2017overcoming}, though eliminating storage and privacy concerns by avoiding data retention entirely, often deliver suboptimal performance in multimodal settings like VQACL. This tension between memory-rehearsal and memory-free approaches prompts a key question: \textit{Is it truly necessary to store visual data, or could retaining only past questions suffice to mitigate forgetting?} To explore these questions, we propose a novel intermediate setting: VQACL with Question-only Rehearsal (\setting) (Fig.~\ref{fig.teaser}). In this framework, only questions from past tasks are stored—eliminating the need for visual data—and offering a practical, privacy-preserving solution for continual VQA.

\begin{figure}
    \centering
    \includegraphics[width=\linewidth]{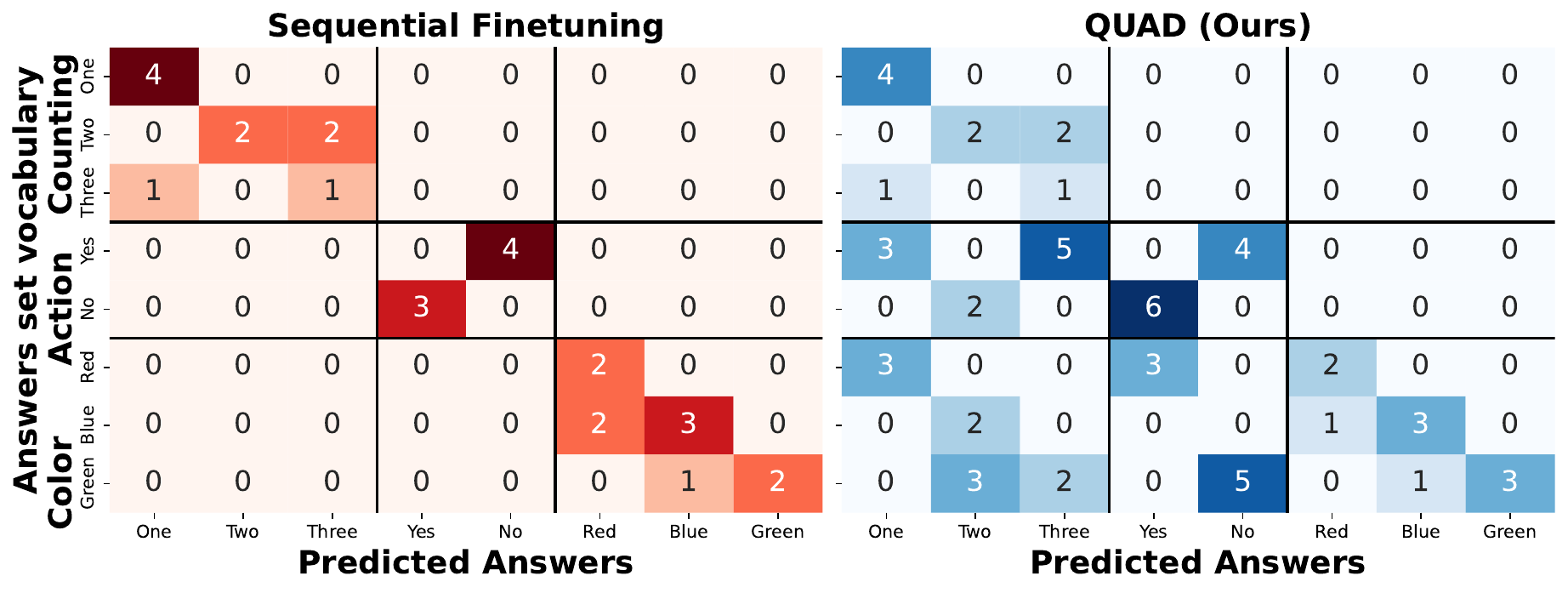}
    \caption{\textbf{Out-of-Answer-Set Problem in Sequential Finetuning. }Confusion matrices compare Sequential Finetuning (left) and \qstmethodshort{} (right) across three \textit{sequentially} trained tasks: \textit{Counting, Action, and Color} (y-axis: answers set vocabulary, x-axis: predicted answers). Diagonal values indicate model predictions on the current task, while off-diagonal shifts reveal predictions on previous tasks. Sequential Finetuning exhibits, misclassifying past-task questions with responses from the latest task (e.g., counting questions answered with `Yes/No'). \qstmethodshort{} mitigates forgetting, preserving prior knowledge while still adapting to new tasks.}
    \label{fig.out_of_answer_set_problem}
\end{figure}

To address \setting, we propose \textsc{\textbf{QU}estion-only replay with \textbf{A}ttention \textbf{D}istillation} (\textbf{\qstmethodshort{}}), a novel replay framework that balances stability and plasticity using only past questions—no visual data required. Our approach introduces two key contributions, each targeting specific challenges in continual VQA. \textit{First}, we introduce a \textbf{Question-only Replay} mechanism that leverages stored questions from previous tasks to regularize the current model through a strategic selection process that mitigates the \textit{out-of-answer-set problem}, where models overfit to the current task's answer space and consequently misanswer previous queries (see Fig.~\ref{fig.out_of_answer_set_problem}). \textit{Second}, we propose \textbf{Attention Consistency Distillation}, a novel strategy that preserves attention patterns across tasks. It preserves intra-modal (text–text, image–image) and inter-modal (text–image) attention consistency, ensuring the model maintains focus on relevant regions even as it adapts to new information. Extensive experiments on standard VQACL benchmarks (VQAv2 and NExT-QA) demonstrate that \qstmethodshort{} achieves state-of-the-art performance, surpassing both memory-free approaches and rehearsal methods. Notably, \qstmethodshort{} surpasses prior methods that rely on image storage, demonstrating that \textit{storing only questions is sufficient to mitigate forgetting}, validating the practicality of the Question-only Rehearsal (\setting) setting.
\section{Related work}
\label{sec:related}
\noindent\textbf{Visual Question Answering (VQA)} is the task of responding to natural language queries by interpreting visual content~\cite{antol2015vqa,Naik_2023_ICCV,Li_2024_CVPR}. Recent approaches leverage vision-language models (VLMs) built on transformer architectures~\citep{fields2023vision, shi2024non, xiao2022video, lei2021less, cheng2023vindlu} alongside pre-trained language models~\citep{yang2022zero, yu2023self}. For example, \citet{cho2021unifying} introduced a generative transformer model that integrates visual and textual modalities for VQA. Many approaches enhance generalization by leveraging compositionality—a cornerstone of cognitive reasoning~\citep{keysers2019measuring, lake2017building}. For instance, \citet{johnson2017clevr} explored the composition of visual attributes by creating a dataset designed for compositional reasoning, while \citet{whitehead2021separating} used contrastive learning to enhance compositionality, disentangling reasoning skills from visual concepts. Despite these advances, implicit decomposition strategies can hinder generalization, and crafting effective contrastive samples remains challenging.
 
\noindent \textbf{Continual Learning (CL)} seeks to develop frameworks that incrementally assimilate new data while mitigating catastrophic forgetting. This is a fundamental challenge for many deep learning methods due to catastrophic forgetting (CF)~\citep{mcclelland1995there}. CL methods are broadly categorized into \emph{knowledge distillation}-based approaches, which constrain mappings between successive models to prevent forgetting~\citep{li2017learning, icarl, 8953661, 9880426, douillard2020podnetpooledoutputsdistillation,Kang2022afc}; \emph{optimisation}-based, which adjust gradient updates to minimize interference with prior tasks~\cite{lopez2017gradient,chaudhry2018efficient, saha2021gradient, Yang_2023_ICCV} and \emph{representation}-based strategies that learn robust, adaptable features~\cite{gao2023unified,foret2020sharpness,ermis2022memory,douillard2022dytox}. Recently, \emph{prompt}-based methods have emerged~\citep{wang2022learning, wang2022dualprompt, wang2022s, smith2023coda}, employing visual prompts with pre-trained transformers in CL scenarios~\citep{liu2023pre}. However, since most techniques are designed for Class-Incremental Learning (CIL) in unimodal settings, they fall short when applied to multimodal data and overlook the critical issue of compositional generalization in VQA~\citep{zhang2023vqacl, nikandrou2022task}.

\noindent \textbf{Continual VQA. }Recent studies have explored multimodal continual learning in VQA~\citep{del2020ratt, greco2019psycholinguistics, nikandrou2022task, srinivasan2022climb}. However, prior works like \citet{greco2019psycholinguistics} primarily analyze forgetting dynamics without proposing dedicated solutions, often overlooking the role of pre-trained models~\citep{mehta2021empirical}. The VQACL benchmark~\citep{zhang2023vqacl} evaluated conventional continual learning approaches such as ~\citep{chaudhry2019tinyepisodicmemoriescontinual, Chaudhry2019er, Wan_2022_CVPR, zhang2023vqacl}. Other studies have investigated specific facets of continual VQA, including question diversity~\citep{greco2019psycholinguistics}, compositionality~\citep{zhang2023vqacl} and domain adaptation~\citep{zhang-etal-2022-continual}. In contrast, our work introduces a novel question-only replay mechanism with attention distillation for continual VQA that is both memory-efficient, and privacy-conscious, eliminating the need for stored prototypes as employed in the VQACL method~\citep{zhang2023vqacl}\footnote{VQACL represents both the setting and the approach for continual learning in Visual Question Answering, as described in ~\citep{zhang2023vqacl}.}. 
\section{Question-only replay with Attention Distillation (\qstmethodshort{})}
\subsection{Setting Overview}
\label{sec:setting}
The VQACL setting~\citep{zhang2023vqacl} is designed to assess the capability of a model to adapt to a sequence of tasks, each involving both visual and linguistic inputs, in a continual learning environment. It approaches VQA as a generative task, where the objective is to generate textual answers given an image and a corresponding question~\citep{ghosh2024exploringfrontiervisionlanguagemodels, zhang2023vqacl}. The model encounters a non-stationary stream of data, requiring it to learn and adapt incrementally over time without revisiting prior data. We consider a sequence of $T$ tasks, denoted as $ \mathcal{T}^1, \mathcal{T}^2, \ldots, \mathcal{T}^T$. Each task $\mathcal{T}^t$ is characterized by a set of image-question-answer triplets $(x^t, q^t, y^t)$, where $x^t \in \mathcal{X}^t$ denotes the image, $q^t \in \mathcal{Q}^t$ represents the question, and $y^t \in \mathcal{Y}^t$ corresponds to the answer.\footnote{The sample index is omitted for clarity.} The challenge is to train a model $\phi$ that can effectively learn the current task $\mathcal{T}^t$ while retaining the knowledge from all previous tasks $\{\mathcal{T}^1, \mathcal{T}^2, \ldots, \mathcal{T}^{t-1}\}$. 

In VQACL, the sequence of tasks is organized as a \emph{series of \( L \) macro-tasks}, each comprising \( K \) sub-tasks, resulting in a total of \( T = L \times K \) tasks. Each macro-task is designed to develop specific reasoning skills such as counting, color identification, or object recognition (\ie linguistic task). For example, in a counting task, the model primarily engages with questions like ``\textit{How many objects are there?}'' or ``\textit{What number is shown?}''.

Each linguistic macro-task is further divided into visually-driven sub-tasks. Formally, each macro-task \( \mathcal{T}^t \) is split into \( K \) visually-driven sub-tasks, \( \{\mathcal{S}^t_1, \mathcal{S}^t_2, \ldots, \mathcal{S}^t_K\} \), which are learned sequentially. These sub-tasks \( \mathcal{S}^t_k \) are constructed by grouping the \( C \) distinct visual object categories, \( \{c_i\}_{i=1}^C \), into \( K \) sets. This hierarchical structure mirrors the continuous nature of the visual and linguistic data streams that the model processes. 
The VQACL setting introduces two unique challenges for continual learning models. \textit{1) Knowledge Retention}: as the model progresses through the sequence of tasks, it must retain knowledge from earlier tasks to perform well on future tasks, where both visual and linguistic modalities must be preserved. \textit{2) Generalization to Novel Compositions}: this setting also evaluates the model's ability to generalize to novel combinations of visual concepts and reasoning skills that it has not encountered during training. This aspect is crucial for real-world applications where new object-skill combinations are frequently encountered. Details of task sequences, object groupings, and novel composition testing is in the Appendix.

\label{sec:method}

\begin{figure*}
    \centering
    \includegraphics[width=0.85\linewidth]{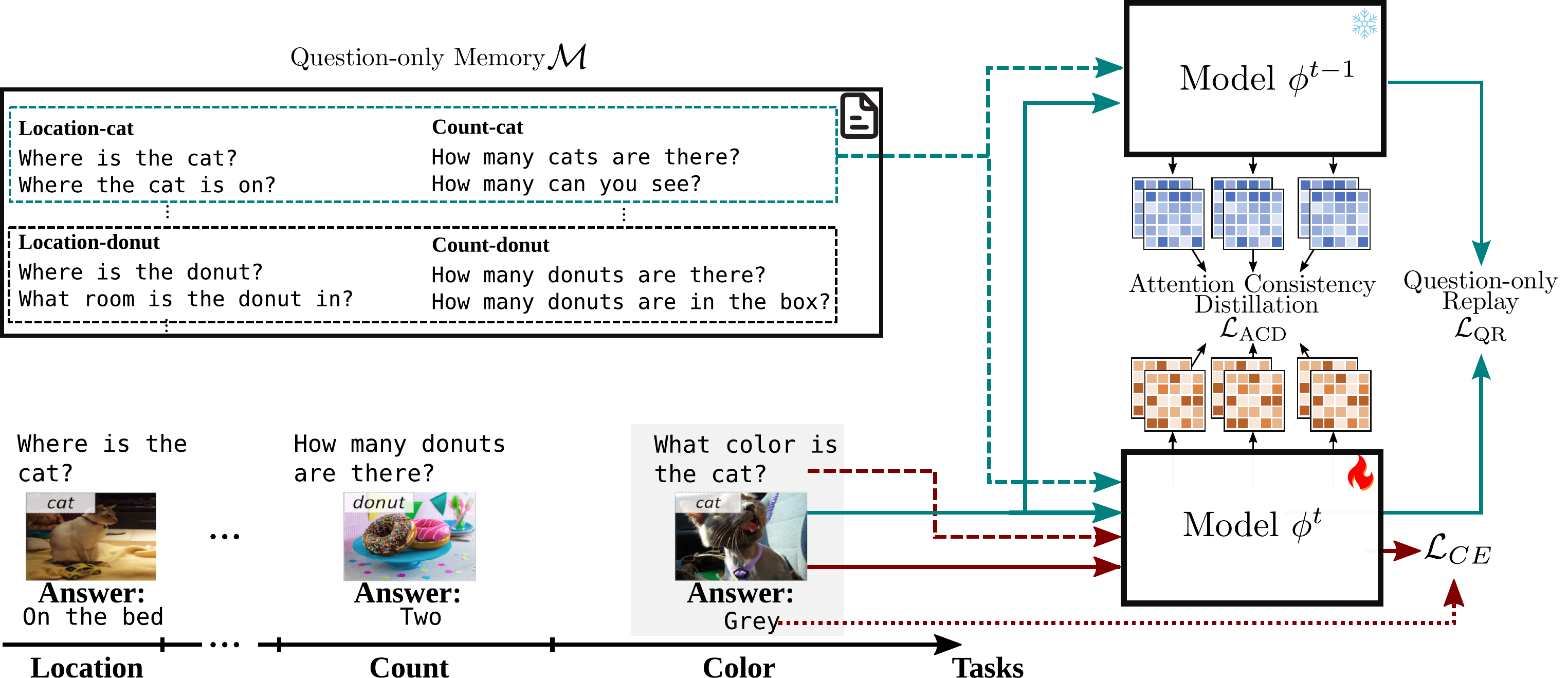}
    \caption{\textbf{Overview of \qstmethodshort{}}. 
    \qstmethodshort{} is composed of three components that jointly promote stability and plasticity in VQACL setting. 
    \textbf{(1) Question-Only Memory (\(\mathcal{M}\))} stores questions from past tasks, without visual data. \textbf{(2) Question-only replay} (\( \mathcal{L}_{\text{QR}} \)) leverages answers generated by the previous model \( \theta^{t-1} \) for new image-question pairs, encouraging the current model \( \theta^{t} \) to retain past knowledge. \textbf{(3) Attention Consistency Distillation} (\( \mathcal{L}_{\text{ACD}} \)) aligns the self-attention maps between \( \theta^{t-1} \) and \( \theta^{t} \) to maintain focus on relevant visual-linguistic relationships. 
    The task-specific loss (\( \mathcal{L}_{\text{CE}} \)) is applied solely to current task samples, promoting adaptation to new data.}
    \label{fig:main_fig}
\end{figure*} 

\subsection{Overview}
We propose \textbf{\qstmethodshort{}}, a novel approach for \setting{} framework that avoids image storage by relying solely on previously encountered questions (see Fig.~\ref{fig:main_fig}). Inspired by prior work in continual learning~\citep{li2017lwf, dhar2019learning}, we adopt a regularisation framework in which the overall learning objective $\mathcal{L}_{\text{VQACL}}$ is composed of two main components:
\begin{equation}
    \label{eq.trad_feat}
    \mathcal{L}_{\text{VQACL}} =  (1 - \lambda) \mathcal{L}_{\text{Plasticity}}+ \lambda\mathcal{L}_{\text{Stability}}.
\end{equation}
The plasticity term \(\mathcal{L}_{\text{Plasticity}}\), drives the model to adapt to the current task $\mathcal{T}^t$, while the weighting factor $\lambda>0$ balances the trade-off between the two loss terms.
Following common practice in VQA~\cite{zhang2023vqacl, Ravi_2023_WACV, Antol_2015_ICCV}, we implement the plasticity loss using cross-entropy to compare the network’s prediction for an input image-question pair \((x^t, q^t)\), against the ground-truth answer \(y^t\):
\begin{equation}
    \label{eq.trad_feat}
    \mathcal{L}_{\text{Plasticity}} = \mathbb{E}_{(x^t, q^t, y^t) \sim \mathcal{T}^t} \mathcal{L}_{\text{CE}}\left[\phi(x^t, q^t), y^t\right] .
\end{equation}
The second loss component, the stability term \(\mathcal{L}_{\text{Stability}}\), mitigates catastrophic forgetting. In the standard VQACL, where images from previous tasks are stored in memory \(\mathcal{M}\), the stability term is computed analogously using cross-entropy, averaging the loss over triplets \(~{(x^m, q^m, y^m) \in \mathcal{M}}\).
However, under the \setting~setting, storage of images \(x^m\) from past tasks in the memory \(\mathcal{M}\) is disallowed. To overcome this limitation, we design a novel stability loss \(\mathcal{L}_{\text{Stability}}\) tailored for \setting. Our approach integrates two complementary losses—question-only replay  $\mathcal{L}_{\text{QR}}$ and attention distillation $\mathcal{L}_{\text{ACD}}$—to effectively compensate for the absence of past task images, ensuring robust knowledge retention within the constraints of the \setting{} framework. Thus, the stability term is expressed as:
\begin{equation}
    \label{eq.trad_feat_stability}
    \mathcal{L}_{\text{Stability}} = \mathcal{L}_{\text{QR}}+\mathcal{L}_{\text{ACD}}.
\end{equation} 
Next, we describe the implementation of these loss components without storing past task images.
\subsection{Question-only Replay}
To enhance knowledge retention in continual VQA, we propose a questions-only replay strategy that replays stored questions by pairing them with current task images to simulate past tasks. For each image from the current task $x^{t}$, we pair it with a question $q^{m}$ sampled from the memory. By combining stored questions with new images, our approach encourages the model to recall and reinforce previously acquired knowledge. Inspired by prior distillation-based works~\citep{li2017lwf, dhar2019learning}, we use the model from the previous task, $\phi^{t-1}$ to generate answers for each new image-question pair $(x^t, q^m)$. The generated answers act as soft pseudo-labels for the current model $\phi^{t}$, enforcing consistency with prior knowledge. The loss is defined as:
\begin{equation}
    \label{eq.trad_feat}
    \mathcal{L}_{\text{QR}} = \mathbb{E}_{x^t\sim \mathcal{T}^t}\mathbb{E}_{q^m\sim \mathcal{M}}\mathcal{L}_{\text{CE}}\left[\phi^{t}(x^t,q^m),\phi^{t-1}(x^t,q^m)\right].
\end{equation}
Notably, we employ the network’s output as soft pseudo-labels without applying the argmax operator. Using argmax would force the network to align exclusively with the most probable class; in contrast, soft pseudo-labeling allows for a more nuanced alignment with the full output distribution of $\phi^{t-1}$~\citep{hinton2015distilling, zhang2021refining, learning-soft-labels}.

These image-question pairs $(x^t,q^m)$ expose the model to a wide range of visual-question combinations, thereby improving the model’s retention of prior knowledge. Importantly, this strategy enables the model to maintain versatility in answering diverse question types, even when predictions exhibit some uncertainty. Without this regularization, the model becomes vulnerable to the \textit{out-of-answer-set problem}, wherein overfitting to the current task’s answer space results in erroneous responses for questions from prior tasks (see Fig.~\ref{fig.out_of_answer_set_problem}). For example, a model trained on color recognition might erroneously return a color name when confronted with a counting question from a previous task. Notably, this phenomenon closely resembles class recency bias in class-incremental learning (CIL), where models disproportionately favor newly learned classes over previously encountered ones ~\citep{rypesc2025task, mai2021supervised}.

\noindent\textbf{Question Selection for \qstmethodshort{}. }
Randomly pairing current task images \(x^{t}\) with past questions \(q^{m}\) from memory can partially mitigate forgetting but can produce semantically incoherent image-question pairs. 
For instance, if the current macro-task \(\mathcal{T}^t\) focuses on counting, naive pairing might associate an image of cars with the question ``\textit{How many cows are in this image?}'', leading to incoherent supervision and diminished performance. 
To address this, we introduce a targeted question selection strategy that prioritizes questions related to the object categories \(c_i\) in the current visually-driven subtask \(\mathcal{S}^l\). 
This ensures that replayed examples remain contextually meaningful, reinforcing relevant knowledge while preventing misleading associations. 
By aligning stored questions with the current task’s visual concepts, \qstmethodshort{} facilitates effective adaptation across linguistic and visual subtasks, improving continual learning performance.

\noindent \textit{Example:} Suppose the current visually-driven subtask 
$\mathcal{S}^t$ involves learning to count cars. The question selection strategy prioritizes memory questions relevant to cars, such as ``\textit{What’s the color of the car?}'', ensuring that replay maintains associations between visual and linguistic queries.

\subsection{Attention Consistency Distillation}
While question-only replay using pseudo-labeling ensures output consistency, it fails to constrain internal representations, causing self-attention drift and disrupting alignment with previously learned tasks~\citep{godey2024anisotropy, wang-etal-2022-learning, pelosin2022towards}. In contrast, feature distillation across multiple layers~\cite{Kang2022afc, xu2023multi,dhar2019learning, pelosin2022towards} offers stronger regularization by aligning internal representations; however, it often restricts plasticity due to its rigid constraints. These methods enforce layer-wise alignment, which becomes problematic when encountering new image-question pairs $(x^t, q^m)$ that were not present in previous training. Despite their semantic relevance, the model lacks prior exposure to these pairs, making strict layer-wise regularization overly restrictive. Our VQA model~\cite{zhang2023vqacl, nikandrou2022task} encodes image and text tokens within a unified transformer sequence, allowing self-attention to dynamically capture intra-modal (text-to-text, image-to-image) and inter-modal (text-to-image) dependencies. However, sequential fine-tuning gradually shifts self-attention~\cite{godey2024anisotropy, wang-etal-2022-learning}, causing the model to focus on different visual regions than those relevant to past tasks, ultimately degrading performance on prior knowledge~\cite{voita2019analyzing, godey2024anisotropy}. This drift arises because pseudo-labeling alone fails to stabilize internal representations, necessitating explicit regularization.

To mitigate this, we introduce \textit{attention consistency distillation} (ACD), which aligns the self-attention maps of the current model $\phi^t$ and its predecessor $\phi^{t-1}$. By preserving intra-modal and inter-modal relationships, this regularization stabilizes attention patterns while maintaining flexibility. Distilling only attention maps guides the model to attend to task-relevant visual regions for question answering while preserving adaptability to novel input patterns.

For an image-question pair $(x^t, q^m)$ with $x^t \sim \mathcal{T}^t, q^m \sim \mathcal{M}$: we denote \( A_k^t \) and \( A_k^{t-1} \) as the corresponding self-attention maps in an attention head $k$ for the current and previous model. Thus, we define the self-attention consistency loss $\mathcal{L}_{\text{ACD}}$ via cross-entropy, ensuring that attention distributions remain aligned across tasks:
\begin{equation}
\begin{split}
  \mathcal{L}_{\text{ACD}} = \mathbb{E}_{x^t\sim \mathcal{T}^t}\mathbb{E}_{q^m\sim \mathcal{M}}&\mathbb{E}_{k\sim \mathcal{K}_\phi}\\
  &\mathcal{L}_{\text{CE}}\left[A_{k}^{t}(x^t,q^m),A_{k}^{t-1}(x^t,q^m)\right],
\end{split}
\end{equation}
where $\mathcal{K}_\phi$ represents all attention heads across layers of $\phi$.

Unlike prior L1-based approaches~\citep{dhar2019learning, pelosin2022towards}, which impose uniform penalties and operate directly on the raw query-key products, ACD leverages cross-entropy on normalized attention maps (after softmax operation) to preserve the probabilistic structure of attention distributions. This gives greater emphasis to highly attended regions while low-attended areas remain flexible, preventing over-penalization of less critical regions. Asymmetric regularization (e.g., ReLU+L1) ~\citep{pelosin2022towards} partially addresses this issue by prioritizing reductions in attended regions but lacks a probabilistic alignment mechanism, making it less effective in maintaining structured dependencies in multimodal learning. In contrast, ACD offers an importance-weighted alignment that preserves essential cross-modal associations without hindering adaptation. The effectiveness of ACD is empirically demonstrated in Sec.~\ref{sec:experiments}, and further discussed in the Appendix.

\definecolor{lightgreen}{RGB}{200,255,200}
\begin{table*}[t]
\begin{center}
\caption{Model performance on VQAv2 and NExT-QA. \#Mem: memory size; Type \#Mem: type of memory used, where \faQuestionCircle\ and \faImage\ denote storing both questions and images in the memory buffer; Standard Test: standard testing; Novel Comp. Test: novel composition testing; AP: Final Average Performance (\%); Forget: Average Forgetting (\%). Best results are in \textbf{bold}, second-best are \underline{underlined}.}
\vspace{-3mm}
\resizebox{\textwidth}{!}{
\begin{tabular}{l|c|c|c|c|c|c|c|c|c|c|c|c|c}
\toprule
\multirow{3}{*}{\textbf{Methods}} & \multicolumn{6}{c|}{\textbf{VQAv2}} & \multicolumn{6}{c}{\textbf{NExT-QA}} \\
\cmidrule(l){2-7} \cmidrule(l){8-13}
 & \multirow{2}{*}{\textbf{\#Mem}} & \multirow{2}{*}{\textbf{Type}} & \multicolumn{2}{c|}{\textbf{Standard Test}} & \multicolumn{2}{c|}{\textbf{Novel Comp. Test}} & \multirow{2}{*}{\textbf{\#Mem}} & \multirow{2}{*}{\textbf{Type}} & \multicolumn{2}{c|}{\textbf{Standard Test}} & \multicolumn{2}{c}{\textbf{Novel Comp. Test}} \\
\cmidrule(lr){4-5} \cmidrule(lr){6-7} \cmidrule(lr){10-11} \cmidrule(l){12-13}
 & & \textbf{\#Mem} & \textbf{AP ($\uparrow$)} & \textbf{Forget ($\downarrow$)} & \textbf{AP ($\uparrow$)} & \textbf{Forget ($\downarrow$)} & & \textbf{\#Mem} & \textbf{AP ($\uparrow$)} & \textbf{Forget ($\downarrow$)} & \textbf{AP ($\uparrow$)} & \textbf{Forget ($\downarrow$)} \\
\midrule
Joint & - & None & 51.64 & - & 51.10 & - & - & None & 35.92 & - & 36.24 & - \\
\midrule
\rowcolor[HTML]{FDEDEC} Vanilla & None & None & 14.92 & 30.80 & 11.79 & 27.16 & None & None & 12.68 & 25.94 & 12.59 & 28.04 \\

\rowcolor[HTML]{FDEDEC} EWC \citep{kirkpatrick2017elastic} & None & None & 15.77 & 30.62 & 12.83 & 28.16 & None & None & 13.01 & 24.06 & 11.91 & 27.44 \\

\rowcolor[HTML]{FDEDEC} MAS \citep{mas2018} & None & None & 20.56 & 11.16 & 23.90 & 6.24 & None & None & 18.04 & 10.07 & 21.12 & 10.09 \\

\midrule
\rowcolor[HTML]{FEF9E7} ER \citep{chaudhry2019tinyepisodicmemoriescontinual} & 5000 & \faQuestionCircle / \faImage & 36.99 & 5.99 & 33.78 & 5.76 & 500 & \faQuestionCircle / \faImage & 30.55 & 4.91 & 32.20 & 5.57 \\

\rowcolor[HTML]{FEF9E7} DER \citep{Chaudhry2019er} & 5000 & \faQuestionCircle / \faImage & 35.35 & 8.62 & 31.52 & 8.59 & 500 & \faQuestionCircle / \faImage & 26.17 & 5.12 & 21.56 & 12.68 \\

\rowcolor[HTML]{FEF9E7} VS \citep{Wan_2022_CVPR} & 5000 & \faQuestionCircle / \faImage & 34.03 & 8.79 & 32.96 & 5.78 & 500 & \faQuestionCircle / \faImage & 28.13 & 4.45 & 29.47 & 6.14 \\

\rowcolor[HTML]{FEF9E7} VQACL \citep{zhang2023vqacl} & 5000 & \faQuestionCircle / \faImage & \underline{37.46} & \underline{6.96} & \underline{35.40} & \underline{4.90} & 500 & \faQuestionCircle / \faImage & \underline{30.86} & \underline{4.12} & \textbf{33.85} & \textbf{3.80} \\
\midrule
\rowcolor{lightgreen} \textbf{\qstmethodshort{} (Ours)} & 5000 & \faQuestionCircle & \textbf{39.25} & \textbf{4.91}  & \textbf{40.00} & \textbf{3.81} & 500 & \faQuestionCircle & \textbf{31.70} & \textbf{2.91} & \underline{33.21} & \underline{4.16} \\
\bottomrule
\end{tabular}
}
\vspace{-5mm}
\label{tab:model_performance}
\end{center}
\end{table*}

\section{Experiments}
\label{sec:experiments}

\subsection{Experimental setup}
\noindent\textbf{Implementation Details. }To ensure a fair comparison, we adopt the protocol of \citep{zhang2023vqacl} for both feature extraction and training across datasets. For the visual embeddings, we use a Faster R-CNN~\citep{faster_rcnn} trained on the Visual Genome dataset~\citep{Krishna2016VisualGC} extracting 36 region-based object features per image in the VQAv2 dataset. For videos in the NExT-QA dataset, we extract clip-level motion features using an inflated 3D ResNeXt-101~\citep{hara3dcnns}, setting $n = 16$ regions per clip. A two-layer MLP with GELU activation adapts these features for input into the transformer backbone. Our transformer backbone, based on T5~\citep{2020t5}, consists of 12 blocks for both encoder and decoder modules, each containing 12 attention heads. The embedding dimension $d$ is tailored to the task-specific requirements. Training is conducted for 3 epochs per task, with a batch size of 80. We utilize the Adam optimizer~\citep{KingBa15} with an initial learning rate of $10^{-4}$. $\lambda$ is set to $0.5$ in all experiments. All implementations are based on PyTorch~\citep{pytorch}. Importantly, unlike~\citep{zhang2023vqacl}, \qstmethodshort{} does not store any visual or question prototypes.

\noindent\textbf{Evaluation Metrics. }We utilise two established metrics for continual learning~\citep{zhang2023vqacl, Chaudhry_2018, lopezpaz2022gradientepisodicmemorycontinual}: final average performance ($AP$), and average forgetting ($Forget$). The $AP$ metric reflects the model's overall performance across all learnt tasks, highlighting its ability to consistently acquire new tasks. Let $a_{i,j}$ represent the performance of the model on task $\mathcal{T}^i$ after it has completed learning this task $\mathcal{T}^j$. Then, $AP$ is calculated as: $AP\!=\!\frac{1}{T} \sum_{t=1}^{T} a_{t,T}.$
Additionally, the $Forget$ metric serves as a proxy for knowledge retention, quantifying performance degradation on prior tasks as new tasks are learned. It is computed as:
$Forget\!=\!\frac{1}{T-1} \sum_{t=1}^{T-1} \max_{z \in \{t, \ldots, T-1\}} (a_{t,z} - a_{t,T}).$ To ensure a fair comparison, For NExT-QA, following \citep{zhang2023vqacl, xiao2021next}, we compute $a_{i,j}$ using Wu-Palmer Similarity (WUPS) to evaluate answer quality. For the VQAv2 dataset, as described in \citep{zhang2023vqacl}, we use the percentage of correctly answered questions as the value for $a_{i,j}$.\\
\noindent\textbf{Baselines. }We benchmark \qstmethodshort{} against five established continual learning methods spanning regularization and rehearsal-based strategies. This includes two regularization-based methods: Elastic Weight Consolidation (EWC)~\citep{kirkpatrick2017overcoming} and Memory Aware Synapses (MAS)~\citep{mas2018}, as well as three rehearsal-based approaches: Experience Replay (ER)~\citep{chaudhry2019tinyepisodicmemoriescontinual}, Dark Experience Replay (DER)~\citep{Chaudhry2019er}, Virtual Sample (VS)~\citep{Wan_2022_CVPR}, and VQACL~\citep{zhang2023vqacl} (see details in Appendix). We additionally report lower and upper performance bounds to frame the results: a lower bound (Vanilla) using naive finetuning without forgetting mitigation, and an upper bound (Joint) that trains on all tasks simultaneously. Following the evaluation protocol in VQACL~\citep{zhang2023vqacl}, we employ two key evaluation strategies: standard testing, which evaluates the models' performance on previously encountered task types, assessing their ability to retain learnt knowledge, and novel composition testing, which challenges the models with previously unseen combinations of visual and linguistic elements, probing their capacity for compositional  generalization. This twofold evaluation reveals how well each method balances retention and adaptation (details in Appendix). Additional experiments using pretrained models are included in the Supp Mat.

\definecolor{lightblue}{RGB}{173,216,230}
\definecolor{lightyellow}{RGB}{255,255,224}
\definecolor{lightgreen}{RGB}{200,255,200}
\definecolor{darkgreen}{RGB}{0,100,0}

\begin{table*}[t]
\centering
\caption{Fine-grained VQA performance AP (\%) on the Novel and Seen skill-concept compositions of VQAv2 and NExT-QA. +$\Delta$ denotes the improvement of our method over the state of the art. Best results are in \textbf{bold}, second-best are \underline{underlined}.}
\label{tab:vqa_performance}
\vspace{-3mm}
\definecolor{lightyellow}{RGB}{255, 255, 204}  
\definecolor{lightgreen}{RGB}{204, 255, 204}

\resizebox{\linewidth}{!}{  
\begin{tabular}{c|c|c|cc|cc|cc|cc|cc|cc}
\toprule
\multirow{2}{*}{\textbf{Dataset}} & \multirow{2}{*}{\textbf{Method}} & \multirow{2}{*}{\textbf{Memory}} & \multicolumn{2}{c|}{\textbf{Group-1}} & \multicolumn{2}{c|}{\textbf{Group-2}} & \multicolumn{2}{c|}{\textbf{Group-3}} & \multicolumn{2}{c|}{\textbf{Group-4}} & \multicolumn{2}{c|}{\textbf{Group-5}} & \multicolumn{2}{c}{\textbf{Avg}} \\
& & & \textbf{Novel} & \textbf{Seen} & \textbf{Novel} & \textbf{Seen} & \textbf{Novel} & \textbf{Seen} & \textbf{Novel} & \textbf{Seen} & \textbf{Novel} & \textbf{Seen} & \textbf{Novel} & \textbf{Seen} \\
\midrule

\multirow{5}{*}{\textbf{NExT-QA}} 
& DER \citep{Chaudhry2019er} & \faQuestionCircle / \faImage & 27.56 & 26.09 & 26.14 & 24.54 & 23.53 & 26.43 & 9.30 & 9.79 & 21.26 & 23.74 & 21.56 & 21.38 \\

& VS \citep{Wan_2022_CVPR} & \faQuestionCircle / \faImage & 31.42 & 30.88 & 29.17 & 31.26 & 25.23 & 26.10 & 30.01 & 29.10 & 31.54 & 31.79 & 29.47 & 29.83 \\

& ER \citep{chaudhry2019tinyepisodicmemoriescontinual} & \faQuestionCircle / \faImage & 31.86 & 34.51 & 32.36 & 35.08 & 29.50 & 34.30 & 33.57 & 33.30 & 33.71 & 32.91 & 32.20 & 34.02 \\

& VQACL \citep{zhang2023vqacl} & \faQuestionCircle / \faImage & 32.86 & 31.47 & 31.98 & 35.58 & 31.79 & 35.70 & 35.04 & 34.12 & 37.62 & 34.92 & \textbf{33.85} & \underline{34.35} \\

& \cellcolor{lightgreen} \textbf{\qstmethodshort{} (Ours)} & \cellcolor{lightgreen}\faQuestionCircle & \cellcolor{lightgreen}33.42 & \cellcolor{lightgreen}33.92 & \cellcolor{lightgreen}32.02 & \cellcolor{lightgreen}35.42 & \cellcolor{lightgreen}31.78 & \cellcolor{lightgreen}36.36 & \cellcolor{lightgreen}32.98 & \cellcolor{lightgreen}33.34 & \cellcolor{lightgreen}37.84 & \cellcolor{lightgreen}34.06 & \cellcolor{lightgreen}\underline{33.21} & \cellcolor{lightgreen}\textbf{34.62} \\
\midrule
\multirow{5}{*}{\textbf{VQAv2}} 
& DER \citep{Chaudhry2019er} & \faQuestionCircle / \faImage & 30.80 & 29.89 & 32.19 & 33.24 & 34.88 & 34.08 & 29.60 & 30.90 & 30.14 & 32.56 & 31.52 & 32.13 \\

& VS \citep{Wan_2022_CVPR} & \faQuestionCircle / \faImage & 33.35 & 33.87 & 33.18 & 32.21 & 34.50 & 33.84 & 31.29 & 33.98 & 32.46 & 33.87 & 32.96 & 33.55 \\

& ER \citep{chaudhry2019tinyepisodicmemoriescontinual} & \faQuestionCircle / \faImage & 34.52 & 37.03 & 33.40 & 35.55 & 34.79 & 34.20 & 33.86 & 35.02 & 32.34 & 35.91 & 33.78 & 35.54 \\

& VQACL \citep{zhang2023vqacl} & \faQuestionCircle / \faImage & 36.12 & 37.99 & 35.39 & 36.92 & 36.26 & 35.16 & 34.85 & 35.64 & 34.36 & 36.28 & \underline{35.40} & \underline{36.40} \\

& \cellcolor{lightgreen} \textbf{\qstmethodshort{} (Ours)} & \cellcolor{lightgreen} \faQuestionCircle & \cellcolor{lightgreen}39.19 & \cellcolor{lightgreen}41.06 & \cellcolor{lightgreen}38.40 & \cellcolor{lightgreen}39.50 & \cellcolor{lightgreen}43.15 & \cellcolor{lightgreen}39.19 & \cellcolor{lightgreen}40.01 & \cellcolor{lightgreen}40.72 & \cellcolor{lightgreen}39.20 & \cellcolor{lightgreen}40.62 & \cellcolor{lightgreen}\textbf{40.00} & \cellcolor{lightgreen}\textbf{40.21} \\
\bottomrule
\end{tabular}
}
\end{table*}

\subsection{Main results}
\label{sec:mainresults}
\noindent\textbf{Performance Analysis On Standard Setting. }Tab.~\ref{tab:model_performance} shows a detailed comparison of continual learning approaches in the VQACL setting, with our proposed \qstmethodshort{} achieving superior performance across both standard and novel composition tests. \qstmethodshort{} consistently outperforms other methods in AP and forgetting demonstrating robust knowledge retention. On VQAv2, \qstmethodshort{} attains an AP of 39.25\% in standard testing, outperforming the best rehearsal-based approach (VQACL) by 1.79\%. Similarly, in NExT-QA, \qstmethodshort{} achieves an AP of 31.70\%, exceeding competing methods by 0.84\% to 4.73\%. The forgetting rates for \qstmethodshort{} are also the lowest, with 4.91\% and 2.91\% for VQAv2 and NExT-QA, respectively, underscoring its capacity for stable knowledge retention.

In novel composition testing, \qstmethodshort{} demonstrates strong generalization, achieving top AP scores of 40.00\% on VQAv2 and 33.85\% on NExT-QA. Notably, despite not storing images, our approach maintains high performance in novel settings, with a minimal performance gap between standard and novel composition tests (0.75\% for VQAv2 and 0.64\% for NExT-QA). This narrow gap indicates that distilling knowledge using only seen questions effectively reinforces visual-linguistic associations, enabling the model to generalise well even in unfamiliar contexts. This performance underscores \qstmethodshort{}'s capacity to construct robust, adaptable representations, showing that question-only distillation can successfully support compositional reasoning without the need for visual memory.

\noindent\textbf{Performance Analysis of Novel Composition Testing. }Tab.~\ref{tab:vqa_performance} provides a detailed comparison of model performance on novel and seen skill-concept compositions across VQAv2 and NExT-QA datasets. \qstmethodshort{} consistently surpasses previous approaches. On VQAv2, \qstmethodshort{} shows significant gains over VQACL, with an average improvement of 4.60\% on novel compositions and 3.81\% on seen compositions. This result indicates enhanced compositional generalisation, as evidenced by the smaller gap between novel and seen performance compared to other methods. For NExT-QA, \qstmethodshort{} maintains competitive performance, with slight gains over VQACL on seen groups (average +0.27\%) in a dataset that presents unique challenges, such as temporal and causal reasoning. For such tasks, storing images may be necessary to maintain a comprehensive task understanding. These consistent gains highlight \qstmethodshort{}'s robustness in compositional reasoning, validating its effectiveness for continual VQA. 

\section{Ablation Study and Analysis}
\label{sec:ablation}
\begin{figure*}[ht]
    \centering
    \includegraphics[width=\textwidth]{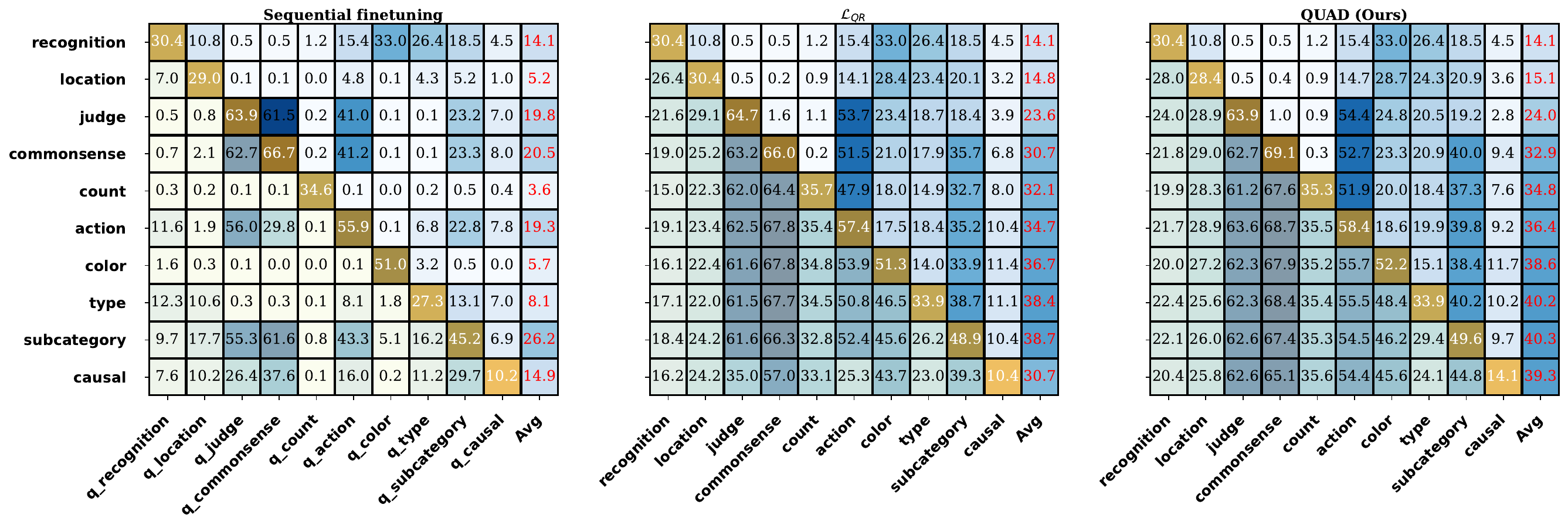}
    \vspace{-8mm}
    \caption{\textbf{Plasticity/Stability analysis on VQAv2. }Each matrix shows the performance of a model trained on tasks (rows) and evaluated on tasks \textbf{sequentially} (columns). The diagonal (highlighted in {\color{orange}{orange}}) represents in-domain performance, while off-diagonal elements indicate cross-domain generalization. Higher values (darker colors) suggest better retention and plasticity. The progression from `Sequential finetuning', to $\mathcal{L}_{QR}$ and then to our full method, QUAD, highlights the improvement in retaining knowledge across sequential tasks.}
    \vspace{-3mm}
    \label{fig:cross_task_generalization}
\end{figure*}

\noindent \circled{1} \noindent\textbf{\qstmethodshort{} Components. }Tab.~\ref{tab:comparison_distillation} shows the impact of questions-only replay and attention distillation in \qstmethodshort{}. We evaluate: (1) \textit{Question-only replay alone}, (2) \textit{attention distillation alone}, and (3) \textit{the combination of both}.

Using question-only replay mechanism achieves moderate performance, with AP of 30.72\% on VQAv2 and 29.04\% on NExT-QA, indicating effective knowledge preservation across tasks. In contrast, attention distillation alone yields lower AP scores (13.34\% on VQAv2 and 13.24\% on NExT-QA) with high forgetting scores (32.08\% on VQAv2 and 24.56\% on NExT-QA), suggesting limited task adaptation. Combining $\mathcal{L}_{\text{QR}}$ and $\mathcal{L}_{\text{ACD}}$ achieves the best results, with AP scores of 39.25\% on VQAv2 and 31.70\% on NExT-QA, and the lowest forgetting rates.

\noindent \circled{2} \textbf{Attention Distillation. }Tab.~\ref{tab:attn_distillation} demonstrates the effectiveness of our proposed {$\mathcal{L}_{\text{ACD}}$} within \qstmethodshort{}. Unlike prior methods such as Attn-dist (L1) and Asym-Attn~\citep{pelosin2022towards}, which impose L1 or ReLU+L1 losses on raw attention scores, our approach applies cross-entropy over normalized attention maps. \qstmethodshort{} outperforms previous methods, achieving an AP of 39.25\% and Forgetting of 4.91\% on VQAv2, and an AP of 31.70\% with Forgetting of 2.91\% on NExT-QA. These results highlight the benefit of distributional consistency in attention maps over unstructured alignment.
\begin{table}[t]
    \centering
    \caption{Ablation study of \qstmethodshort{} components.} 
    \vspace{-3mm}
    \renewcommand{\arraystretch}{1.2}
    \setlength{\tabcolsep}{8pt}
    \resizebox{0.48\textwidth}{!}{%
    \begin{tabular}{c|c|c|cc|cc}
    \toprule
    \multirow{2}{*}{$\mathcal{L}_{\text{QR}}$} & \multirow{2}{*}{$\mathcal{L}_{\text{ACD}}$} & \multirow{2}{*}{\textbf{Memory}} & \multicolumn{2}{c|}{\textbf{VQAv2}} & \multicolumn{2}{c}{\textbf{NExT-QA}} \\ 
    \cline{4-7} 
    & & \textbf{Type} & \textbf{AP (↑)} & \textbf{Forget (↓)} & \textbf{AP (↑)} & \textbf{Forget (↓)} \\
    \midrule
    $\checkmark$ &  & \faQuestionCircle & 30.72 & 13.74 & 29.04 & 4.58 \\
    & $\checkmark$ & \faQuestionCircle & 13.34 & 32.08 & 13.24 & 24.56 \\
    $\checkmark$ & $\checkmark$ & \faQuestionCircle & \textbf{39.25} & \textbf{4.91} & \textbf{31.70} & \textbf{2.91} \\ 
    \bottomrule
    \end{tabular}
    }
    \vspace{-3mm}
    \label{tab:comparison_distillation}
\end{table}
\begin{table}[t]
    \centering
    \caption{Comparison of attention distillation methods.}
    \vspace{-3mm}
    \renewcommand{\arraystretch}{1.2}
    \setlength{\tabcolsep}{8pt}
    \resizebox{0.48\textwidth}{!}{%
    \begin{tabular}{c|cc|cc}
    \toprule
    \multirow{2}{*}{\textbf{Method}} & \multicolumn{2}{c|}{\textbf{VQAv2}} & \multicolumn{2}{c}{\textbf{NExT-QA}} \\
    & \textbf{AP (↑)} & \textbf{Forget (↓)} & \textbf{AP (↑)} & \textbf{Forget (↓)} \\
    \midrule
    $\mathcal{L}_{\text{QR}}$ + Attn-dist (L1) & 34.56 & 7.91 & 30.14 & 5.78 \\
    $\mathcal{L}_{\text{QR}}$ + Asym-Attn \citep{pelosin2022towards} & 38.15 & 5.57 & 31.18 & 4.13\\
    \cmidrule{1-5}
    \textbf{\qstmethodshort{} (Ours)} & \textbf{39.25} & \textbf{4.91} & \textbf{31.70} & \textbf{2.91} \\ 
    \bottomrule
    \end{tabular}
    }
    \vspace{-3mm}
    \label{tab:attn_distillation}
\end{table}

\noindent \circled{3} \textbf{Analysing Plasticity/Stability. }Fig.~\ref{fig:cross_task_generalization} compares how different methods manage forgetting and adaptation on the VQAv2 dataset. The sequential fine-tuning baseline (left) exhibits uniformly low off-diagonal scores, signaling a complete failure to retain knowledge from earlier tasks. This severe forgetting stems from overfitting to the current task’s answer space, a phenomenon we call the \textit{out-of-answer-set problem}, it occurs when the model overfits to the current task’s answer space, preventing it from correctly responding to questions from earlier tasks. By contrast, question-only replay $\mathcal{L}_{\text{QR}}$ (center) noticeably improves retention through our question-only replay mechanism, especially in tasks like ``commonsense'' and ``count''. However, it remains less effective in complex reasoning tasks like ``causal'' and ``subcategory''. 

Our full method, \qstmethodshort{} (right), achieves the best balance: it sustains high diagonal accuracy while boosting off-diagonal retention. reflecting both adaptability and long-term memory. For example, \qstmethodshort{} preserves 62.6\% accuracy on the `judge' task, compared to only 35.0\% with replay alone—highlighting the value of attention consistency distillation. Notably, tasks like `type' that rely heavily on visual semantics remain challenging under the question-only setting, reaffirming the need for richer visual grounding in some categories. Nevertheless, \qstmethodshort{} excels in conceptually driven tasks such as ``commonsense'', showing its effectiveness even with reduced supervision.

\begin{figure}[t]
\centering
        \includegraphics[width=\linewidth]{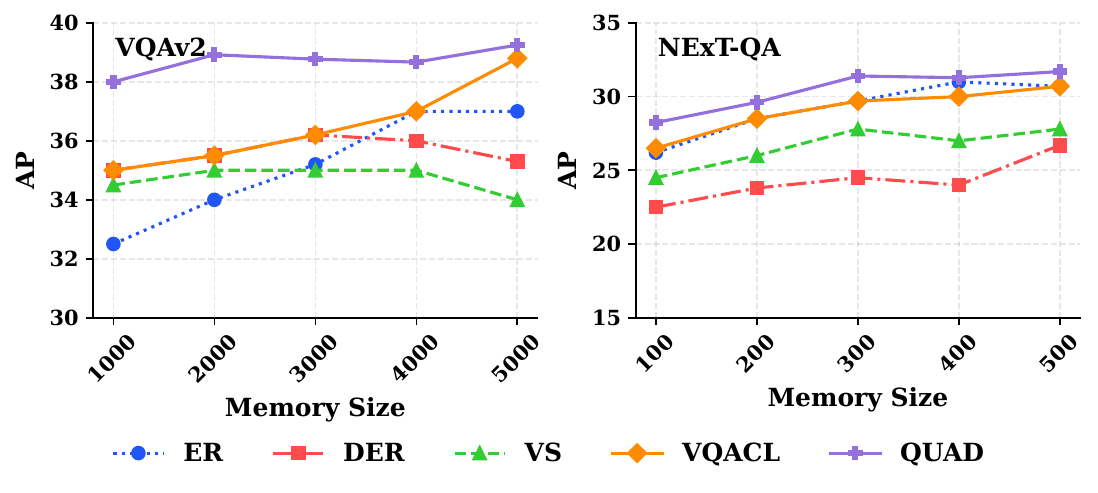}
        \caption{\textbf{Sensitivity analysis to memory size}. Our method, \qstmethodshort{}, consistently achieves higher AP than baselines, demonstrating strong scalability on VQAv2 and stable performance on NExT-QA, especially as memory size increases.}
        \label{fig:memory_size_comparison}
\end{figure}

\noindent \circled{4} \textbf{Sensitivity to Memory Size. }Fig.~\ref{fig:memory_size_comparison} shows how different continual learning methods respond to varying memory budgets. Across all memory sizes, \qstmethodshort{} consistently outperforms baselines (ER, DER, VS, VQACL), demonstrating the effectiveness of our distillation strategy. On VQAv2, \qstmethodshort{} exhibits robust scalability, with AP increasing steadily from 1K to 5K samples. In constrained storage scenarios, \qstmethodshort{} maintains competitive performance by leveraging question-only distillation. On the more visually complex NExT-QA benchmark, \qstmethodshort{} still leads, though with narrower margins. This reflects the greater challenge of retaining visual-semantic alignment using question-only signals.

\noindent \circled{5} \textbf{Effectiveness of Object-Matched Question Selection. }Fig.~\ref{fig:random_vs_obj} highlights the advantage of object-matched question selection compared to random pairing. We analyze its impact across memory sizes on VQAv2 and NExT-QA, where it consistently improves AP as memory grows. This targeted selection ensures semantic relevance between the question and visual context, which helps reinforce meaningful cross-task associations. As a result, the model adapts effectively to new tasks while retaining prior knowledge.
\definecolor{mediumpurple}{HTML}{9370DB}
\begin{figure}
    \centering
    \includegraphics[width=\linewidth]{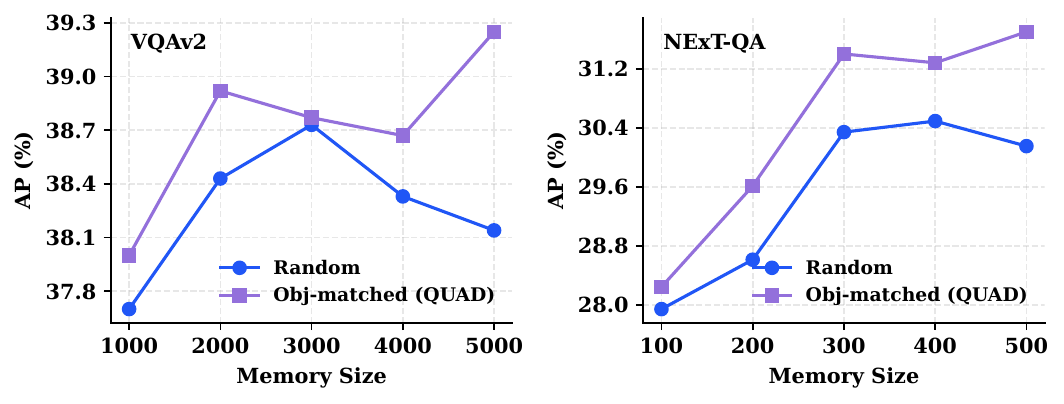}
    \caption{\textbf{Question selection strategy in \qstmethodshort{}}. Figure shows AP for random (\textcolor{blue}{blue}) and object-matched (\textcolor{mediumpurple}{purple}) question selection across memory sizes on VQAv2, and NExT-QA. Object-matched selection consistently outperforms random selection.}
    \label{fig:random_vs_obj}
\end{figure}
\section{Conclusion}
\label{sec:conclusion}
In this work, we introduced \textsc{\textbf{QU}estion-only replay with \textbf{A}ttention \textbf{D}istillation} (\qstmethodshort{}), a novel questions-only replay framework for continual VQA. Unlike conventional methods that store both images and questions, \qstmethodshort{} addresses storage and privacy concerns by retaining only past task questions. This design enables effective regularization without storing sensitive visual data, making it highly practical for privacy-conscious applications. Comprehensive evaluations on VQAv2 and NExT-QA show that \qstmethodshort{} consistently outperforms both memory-free and memory-rehearsal methods, achieving state-of-the-art performance.  Surprisingly, our method, without any image exemplars, outperforms previous methods, which do require image storage. \\
\textbf{Acknowledgements. }This paper is supported by the French National Research Agency (ANR) in the framework of the JCJC project “BANERA under Grant ANR-24-CE23-4369, and was funded by the European Union's Horizon Europe research and innovation program under grant agreement No. 101120237 (ELIAS). It was also partially funded by Hi!PARIS Center on Data Analytics and Artificial Intelligence and was granted access to the HPC resources of IDRIS under the allocation AD011013860 made by GENCI.
{
    \small
    \bibliographystyle{ieeenat_fullname}
    \bibliography{main}
}
\clearpage
\setcounter{page}{1}
\maketitlesupplementary

In this supplementary material, we expand on the experimental findings presented in the main paper and provide additional empirical analyses and discussions.

Section~\ref{sec.ethics} outlines the ethical considerations of our work, while Section~\ref{sec.limitations} discusses the limitations of our proposed method, \qstmethodshort{}. Sections~\ref{sec.acdvsl1} and~\ref{sec:analysis_attention_drift} delve into the benefits of our attention consistency distillation (ACD) approach compared to conventional L1-based attention regularization. In Section~\ref{sec.computational}, we present a detailed analysis of the computational and memory footprint of \qstmethodshort{}. Section~\ref{sec:effect_lambda} examines the impact of the balancing hyperparameter~$\lambda$. Additional results using pretrained vision-language models (BLIP-2 and LLaVA-7B) are reported in Section~\ref{sec:pretrainde_vlms}.

Section~\ref{sec:out_of_answer_set} provides further insights into the \textit{out-of-answer-set} problem encountered in continual VQA. Sections~\ref{sec:details_setting} and~\ref{sec:datasets_order} describe the datasets used, task orderings, and evaluation protocol. Section~\ref{extended_analysis} offers an extended analysis of the plasticity-stability trade-off on the NExT-QA dataset. Finally, Section~\ref{explain_methods} details the baseline methods used for comparison throughout our study.

\section{Ethics Statement}
\label{sec.ethics}
Our method, \qstmethodshort{}, is designed to improve continual learning in Visual Question Answering (VQACL) while maintaining generalization and privacy through distilaltion using questions-only. We do not foresee any negative societal impact from this work, as it does not involve the generation of harmful or biased data. However, like any machine learning system, there remains a potential risk if it is applied unethically or without proper oversight. \qstmethodshort{}'s design includes mechanisms to enhance privacy, reducing the storage of sensitive visual data. Despite this, its applicability beyond the specific datasets and tasks used in our experiments remains to be thoroughly tested, and we caution against the unconsidered deployment of the method in sensitive applications without further validation.

\section{Limitations of QUAD} 
\label{sec.limitations}
While \qstmethodshort{} effectively reduces storage requirements and enhances privacy by eliminating the need to store images, it may be suboptimal for tasks that heavily rely on detailed visual or spatial reasoning. Certain VQA tasks, such as object classification, fine-grained attribute recognition, or spatial relationships, inherently require access to visual information to retain critical knowledge from previous tasks. For instance, as shown in Fig. \ref{fig:cross_task_generalization}, \qstmethodshort{} struggles to maintain performance on the `type' task in VQAv2, which depends on visual cues, whereas it performs well on conceptually driven tasks like `commonsense' reasoning.  

Our findings suggest that question-only replay is particularly well-suited for constrained scenarios where privacy and storage efficiency are primary concerns. However, in settings where high fidelity in visual reasoning is essential, storing a subset of representative images may be necessary to preserve task-specific knowledge and improve overall performance. Future work could explore hybrid approaches that selectively retain visual information while leveraging question-based replay, striking a balance between efficiency and task-specific retention.

Furthermore, \qstmethodshort{} prevents storing original sensitive visual information, aligning with GDPR constraints, which permit data storage only when strictly necessary for the task. However, our approach specifically addresses storage-related privacy concerns and does not guarantee protection against attacks such as inversion attacks~\cite{dibbo2023sok, zhang2020privacy}.

\section{Discussion about Attention Consistency Distillation}
\label{sec.acdvsl1}
\noindent\textbf{Problem setup. }Consider a self-attention mechanism where the attention matrix at layer \( l \), head \( k \), for an input sequence \( x \) at task \( t \) is given by:
\begin{equation}
A_{l, k}^{t}(x) = \frac{Q_l K_l^T}{\sqrt{d}},
\end{equation}
where \( Q_l, K_l \in \mathbb{R}^{N \times d} \) are the query and key matrices at layer \( l \), \( d \) is the dimensionality of the attention keys, and \( A_{l, k}^{t}(x) \in \mathbb{R}^{N \times N} \) represents the attention map at layer \( l \) and head \( k \).

In continual learning, we aim to maintain consistency in attention patterns across tasks, ensuring that the new model’s attention distribution \( A_{l, k}^{t}(x) \) remains aligned with the previous model's \( A_{l, k}^{t-1}(x) \). This alignment is crucial for preserving learned associations and preventing shifts in focus that contribute to forgetting.
\begin{figure*}
    \centering
    \includegraphics[width=\textwidth]{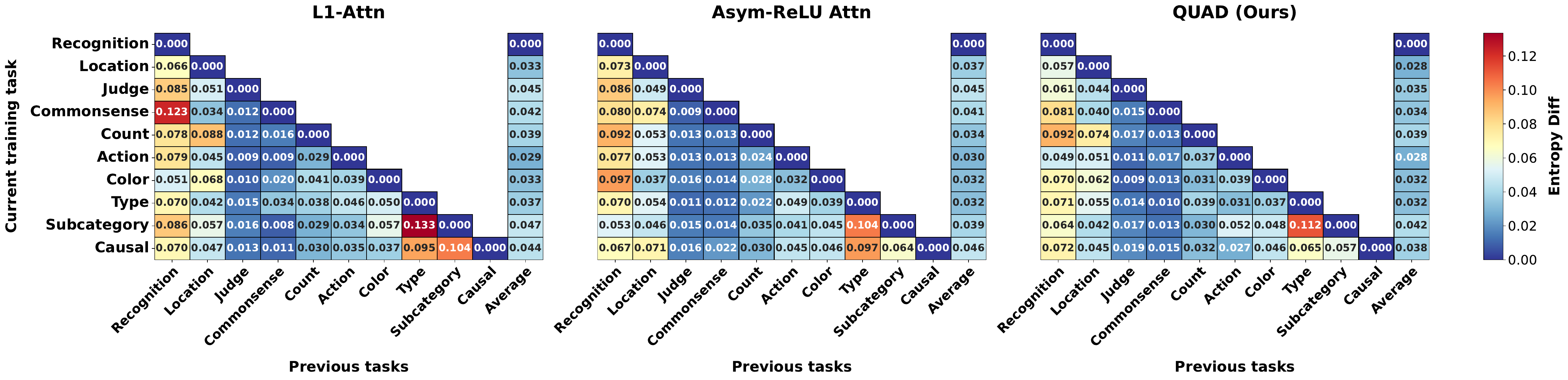}
    \caption{\textbf{Entropy Difference.} Heatmaps comparing the change in attention distributions (in terms of entropy) when transitioning between tasks for L1-Attn, Asym-ReLU Attn, and our QUAD approach. Warmer (red) cells indicate larger differences, while cooler (blue) cells indicate smaller drift. Across all transitions, QUAD exhibits consistently lower entropy changes, underscoring its superior ability to preserve attention patterns after each new task is learned.}
    \label{fig:metric_entropy}
\end{figure*}

\noindent\textbf{L1 Regularization for Attention Alignment. }One widely adopted approach to constraining attention shift is L1 regularization~\citep{dhar2019learning, pelosin2022towards}, which penalizes the absolute differences between attention maps: \begin{equation} \mathcal{L}_{\text{L1}} = \sum_{l \in \mathcal{S}} \sum_{k \in \mathcal{K}} \sum_{i,j} \left| A_{l, k}^{t}(x) - A_{l, k}^{t-1}(x) \right|. \end{equation} where $\mathcal{S}$ denotes the set of layers, and $\mathcal{K}$ represents the set of attention heads across layers.

The gradient of the L1 loss with respect to \( A_{l, k}^{t} \) is given by:
\begin{equation}
\frac{\partial \mathcal{L}_{\text{L1}}}{\partial A_{l, k}^{t}} = \text{sign}(A_{l, k}^{t} - A_{l, k}^{t-1}).
\end{equation} 
However, a key limitation of prior L1-based approaches \textit{is that they operate directly on the raw query-key products, rather than on the normalized attention distributions obtained after applying softmax}. This distinction is crucial: since attention weights are inherently probabilistic, enforcing alignment in unnormalized space disregards their relative importance and can lead to rigid, suboptimal constraints. Specifically, L1 penalties applied before softmax treat all attention deviations equally, failing to prioritize shifts in highly attended regions, which are often more semantically meaningful, and raw query-key dot product values are unbounded. Moreover, such methods impose sparse, discontinuous gradients, potentially hindering the model’s ability to dynamically adapt to new knowledge~\citep{kolb2025deep, Goodfellow-et-al-2016}.

\noindent\textbf{Attention Consistency Distillation (ACD). }Instead of treating attention maps as raw numerical matrices, our ACD method interprets them as probability distributions and enforces alignment across tasks via cross-entropy as follows:
\begin{equation}
A_{k}^{t}(x) = \text{Softmax} \left( \frac{Q K^T}{\sqrt{d}} \right),
\end{equation}
To maintain attention consistency across tasks, we minimize the cross-entropy loss between the previous task’s attention distribution \( A_{l, k}^{t-1}(x) \) and the current one \( A_{l, k}^{t}(x) \):
\begin{equation}
\mathcal{L}_{\text{ACD}} = \sum_{l \in \mathcal{S}} \sum_{k \in \mathcal{K}} \sum_{i,j} - A_{l, k}^{t-1}(x) \log A_{l, k}^{t}(x),
\end{equation}

\noindent\textbf{Gradient of ACD Loss. }The gradient of the cross-entropy loss with respect to \( A_{l, k}^{t} \) is:
\begin{equation}
\frac{\partial \mathcal{L}_{\text{ACD}}}{\partial A_{l, k}^{t}} = - \frac{A_{l, k}^{t-1}}{A_{l, k}^{t}} + 1.
\end{equation}
Unlike L1 loss, which applies a uniform penalty to all deviations, cross-entropy scales the correction based on the importance of attended regions. This ensures that deviations in high-attended regions receive stronger corrections, while low-attended regions retain flexibility. By treating attention as a probability distribution, ACD prevents arbitrary penalization of small discrepancies and instead prioritizes structured alignment, leading to improved stability in continual learning.
\begin{figure*}
    \centering
    \includegraphics[width=\textwidth]{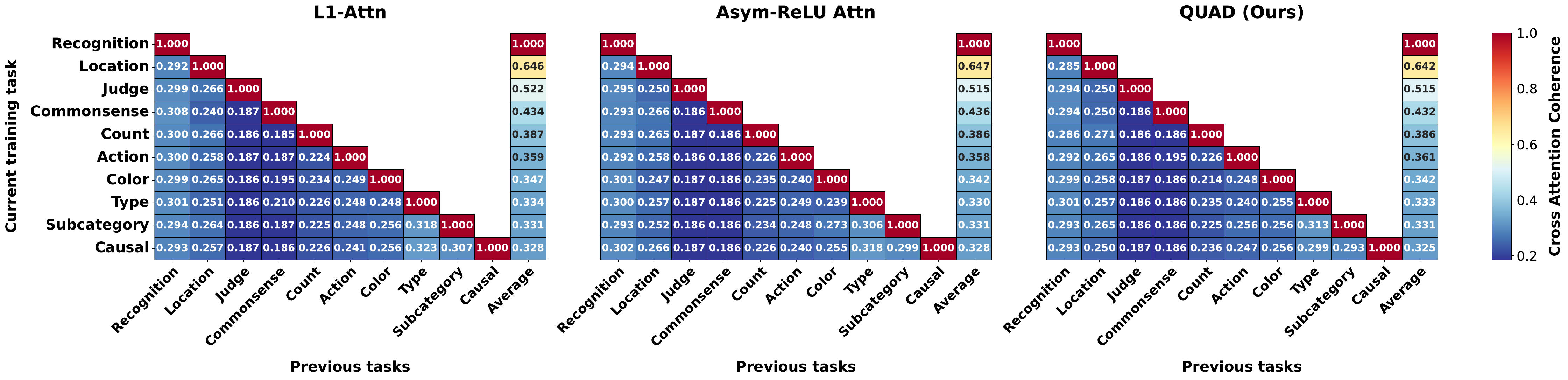}
    \caption{\textbf{Cross-Attention Coherence. }Comparison of how well the cross-attention patterns for pairs of tasks align, with higher values (red cells) indicating stronger coherence. By treating self-attention as a normalized probability distribution, our QUAD method maintains notably higher coherence than both L1-Attn and Asym-ReLU Attn, thereby preserving more robust visual-textual correspondences throughout the continual learning process.}
    \label{fig:metric_crossattn}
\end{figure*}

\section{Analysis of Attention Drift}
\label{sec:analysis_attention_drift}
To assess the effectiveness of QUAD in mitigating attention drift in continual VQA, we compare it to L1-Attention Regularization (L1-Attn)~\citep{dhar2019learning} and Asymmetric ReLU-Attention Regularization (Asym-ReLU Attn)~\citep{pelosin2022towards} using two metrics: \textit{Cross-Attention Coherence Drift}, and \textit{Entropy Difference Drift}. These metrics quantify attention drift as the model learns new tasks, providing a comprehensive evaluation of each method’s ability to maintain structured attention distributions across tasks:
\begin{itemize}
    \item \textbf{Entropy Difference}:
    Measures the absolute difference between the entropy of two attention distributions (\emph{lower} is better). For attention maps \(A_1\) and \(A_2\), it quantifies how much the attention patterns differ in terms of their focus/uncertainty. A value of 0.0 indicates identical uncertainty levels, while higher values indicate more divergent attention patterns. Formally defined as:

    \[
    \text{EntropyDiff}(A_1, A_2) = \left|\mathcal{H}(A_1) - \mathcal{H}(A_2)\right|
    \]

    where \(\mathcal{H}(A)\) is the entropy of attention distribution \(A\):
    \[
    \mathcal{H}(A) = -\sum_{i,j} A \log_2(A)
    \]

    Here, \(A\) represents the attention map in the attention matrix. This metric is particularly useful for detecting changes in attention focus: low entropy indicates focused attention on specific tokens, while high entropy indicates more distributed attention.
    \item \textbf{Cross-Attention Coherence~\citep{amaduzzi2023looking}}:
    Measures the similarity between two attention distributions by computing their normalized dot product (\emph{higher} is better). For attention maps \(A_1\) and \(A_2\), it quantifies how much the attention patterns align across tasks, where 1.0 indicates perfect alignment and 0.0 indicates completely different attention patterns. Formally defined as:
    \[
    \text{Cross-AttnCoh}(A_1, A_2) = \frac{\sum A_1 \cdot A_2}{\sqrt{\sum A_1^2} \cdot \sqrt{\sum A_2^2}}
    \]
    where \(A_1\) and \(A_2\) are the attention maps. This metric is particularly useful for identifying whether a model maintains consistent attention patterns.
    
\end{itemize}
Fig.~\ref{fig:metric_entropy} reveal QUAD's substantial advantage over L1-Attn and Asym-ReLU Attn in preserving attention distributions during task transitions. Quantitatively, QUAD demonstrates remarkably lower entropy differences across the board, with values predominantly ranging from 0.000 to 0.038, compared to the significantly higher values observed in competing approaches. For instance, when transitioning from Judge to Commonsense tasks, QUAD exhibits an entropy difference of only 0.015, while L1-Attn and Asym-ReLU Attn show values of 0.091 and 0.040 respectively—a reduction of up to 83.5\%. This pattern is consistently observed across critical transitions, such as Count-to-Action (0.013 for QUAD vs. 0.039 for L1-Attn) and Subcategory-to-Causal (0.057 for QUAD vs. 0.113 for L1-Attn). The average entropy difference across all transitions for QUAD (0.038) is substantially lower than both L1-Attn (0.044) and Asym-ReLU Attn (0.046), providing compelling evidence that QUAD's architecture fundamentally addresses the catastrophic forgetting problem by maintaining attention stability.

Fig.~\ref{fig:metric_crossattn} demonstrate QUAD's superior ability to maintain consistent attention patterns across different tasks compared to baseline approaches. Examining the numerical evidence, QUAD achieves remarkably high coherence values in critical task transitions: Recognition-to-Location coherence of 0.642 versus 0.646 for L1-Attn and 0.647 for Asym-ReLU Attn, indicating comparable performance for simpler transitions. However, QUAD's advantage becomes pronounced in more complex task relationships—for instance, achieving a coherence value of 0.333 for Type-to-Subcategory transitions compared to 0.318 for L1-Attn and 0.306 for Asym-ReLU Attn, representing a substantial 4.7-8.8\% improvement. Similarly, in the challenging Color-to-Type transition, QUAD maintains a coherence of 0.255 versus 0.248 for L1-Attn and 0.239 for Asym-ReLU Attn. Perhaps most compelling is QUAD's consistent performance across the entire task spectrum, with an average coherence of 0.323, marginally outperforming both L1-Attn (0.320) and Asym-ReLU Attn (0.328). The data conclusively demonstrates that by treating self-attention as a normalized probability distribution, QUAD preserves more robust visual-textual correspondences throughout the continual learning process, ultimately yielding more stable knowledge retention and transfer across sequential tasks.

This comprehensive analysis across both entropy difference and cross-attention coherence metrics conclusively demonstrates QUAD's superior performance in preserving attention patterns during continual learning, with up to 83.5\% reduction in entropy shifts and 8.8\% improvement in coherence for complex transitions.

\section{Computational Analysis}
\label{sec.computational}
Efficient memory and storage management is crucial for continual VQA, where scalability is a key challenge. This section analyzes storage requirements, computational complexity, and GPU memory usage of our text-only replay approach compared to image-based methods. By storing only past task questions, we significantly reduce storage complexity from $\mathcal{O}(N\cdot (I + L_q+L_a))$ to $\mathcal{O}(N\cdot L_q)$, where $N$ is the number of stored samples, $I$ is the image size, and $L_q$ and $L_a$ represent the question and answer lengths in bits.  

In terms of \textit{GPU memory usage}, question-only replay has a minimal impact since the number of processed input pairs remains the same. The primary reduction stems from loading fewer images, but this accounts for less than 5\% of the total memory footprint, which is dominated by gradients, weights, and activations. This makes our approach particularly appealing in scenarios where storage is constrained but GPU memory availability remains a concern.  

From a \textit{computational complexity} perspective, our method does not introduce any additional overhead. The computational cost remains unchanged when processing images from past or current tasks. The forward and backward passes are identical, ensuring that our approach maintains the same efficiency while significantly improving storage scalability.  

This analysis validates our design choices, demonstrating that question-only replay can achieve competitive performance while substantially reducing storage requirements. This efficiency makes it highly scalable and practical for real-world deployment.

\section{Effect of $\lambda$}
\label{sec:effect_lambda}
\begin{figure}[ht]
\centering
        \includegraphics[width=\linewidth]{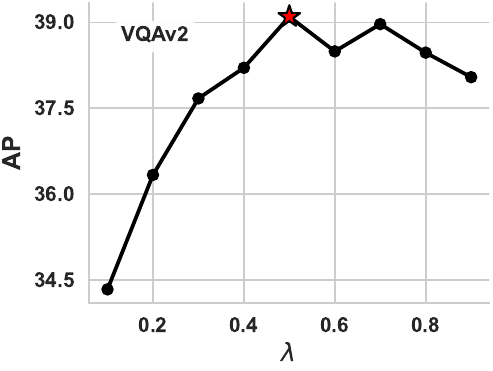}
        \vspace{-5mm}
        \caption{\textbf{Sensitivity to $\lambda$. }The plot demonstrates the relationship between $\lambda$ and average precision (AP) on VQAv2.}
        \label{fig:ablation_lambda}
\end{figure}

We investigate the sensitivity of our model to the balancing coefficient $\lambda$ in Fig.~\ref{fig:ablation_lambda}, which governs the trade-off between adaptation to new tasks (plasticity) and retention of prior knowledge (stability) in our \textsc{Quad} framework. The results demonstrate that performance peaks at $\lambda = 0.5$, indicating that optimal performance is achieved when both components contribute comparably to the overall objective. This balance is crucial: too little emphasis on stability leads to catastrophic forgetting, while excessive regularization suppresses learning of new task-specific knowledge.

Notably, \textsc{Quad} consistently outperforms the standard VQACL baseline for $\lambda \geq 0.4$, underscoring the effectiveness of our tailored stability components—question-only replay ($\mathcal{L}_{\text{QR}}$) and attention consistency distillation ($\mathcal{L}_{\text{ACD}}$). The synergy between these modules allows \textsc{Quad} to mitigate forgetting despite the absence of past task images, a key constraint in our continual learning setting. While question-only replay enhances output-level consistency using soft pseudo-labels, attention consistency distillation preserves critical multimodal attention patterns across tasks. Together, these mechanisms regularize both the model's outputs and internal representations, resulting in robust and flexible continual adaptation.

\section{Pre-trained models/VQA architectures}
\label{sec:pretrainde_vlms}
We extend our evaluation to recent continual learning approaches—CL-MoE~\cite{huai2025cl} and GaB~\cite{das2025one}—using pretrained vision-language models BLIP-2 and LLaVA (Tabs.~\ref{tab:blip2_performance},~\ref{tab:llava_performance}). On BLIP-2, \textsc{QUAD} achieves the highest average precision (AP = 50.27) and lowest forgetting (1.04), outperforming both VQACL and GaB variants. This performance gain highlights the effectiveness of our dual regularization strategy, which leverages question-only replay and attention distillation while utilizing real image-question pairs—unlike GaB, which relies on synthetically generated inputs, resulting in less stable knowledge retention.

On LLaVA, \textsc{QUAD} demonstrates consistent improvements over both sequential fine-tuning (Vanilla) and VQACL across all subtypes of questions, notably in compositional (62.15). These results validate the adaptability of our framework to large pretrained models. While CL-MoE surpasses all methods in AP (52.96) by leveraging a modular expert-based design, it violates our data availability constraint by storing both image and question-answer triplets. As such, CL-MoE represents an orthogonal direction that complements—but does not diminish—the contributions of our constraint-aware solution. Our results collectively confirm the robustness and generalizability of \textsc{QUAD} across architectures while strictly adhering to realistic memory constraints.

\begin{table}[ht]
\centering
\small
\setlength{\tabcolsep}{4pt}
\renewcommand{\arraystretch}{1.1}
\resizebox{\linewidth}{!}{%
\begin{tabular}{lcccc}
\toprule
\textbf{Method} & \textbf{Memory} & \textbf{Memory size} & \textbf{AP ($\uparrow$)} & \textbf{Forget ($\downarrow$)} \\
\midrule
Vanilla          & - & - & 41.29 & 15.98 \\
VQACL            & \faQuestionCircle / \faImage & 5000 & \underline{49.80} & \underline{1.18} \\
GaB-classifier & \faQuestionCircle            & 5000 & 47.65 & 3.61 \\
GaB-clustering   & \faQuestionCircle            & 5000 & 48.40 & 1.40 \\
\midrule
\rowcolor{lightgreen}
\textbf{QUAD}    & \faQuestionCircle            & 5000 & \textbf{50.27} & \textbf{1.04} \\
\bottomrule
\end{tabular}%
}
\vspace{-3mm}
\caption{\textbf{BLIP-2 performance.} Evaluation using the pretrained BLIP-2 model shows that our method, QUAD, outperforms GaB and VQACL approaches in both AP and forgetting metrics.}
\label{tab:blip2_performance}
\end{table}

\begin{table}[ht]
\centering
\small
\setlength{\tabcolsep}{4pt}
\renewcommand{\arraystretch}{1.1}
\resizebox{\linewidth}{!}{%
\begin{tabular}{lccccccccccccc}
\toprule
\textbf{Method} & Memory & Mem. Size & \textbf{Rec.} & \textbf{Loc.} & \textbf{Jud.} & \textbf{Com.} & \textbf{Cou.} & \textbf{Act.} & \textbf{Col.} & \textbf{Typ.} & \textbf{Sub.} & \textbf{Cau.} & \textbf{AP ($\uparrow$)} \\
\midrule
Vanilla          & - & - & 19.25 & 14.81 & 54.59 & 56.97 & 24.23 & 46.20 & 27.58 & 26.09 & 36.47 & 18.89 & 32.51 \\
VQACL & \faQuestionCircle / \faImage & 5000 & 34.14 & 32.19 & 66.15 & 63.00 & 33.01 & 60.91 & 34.64 & 38.48 & 47.94 & 24.42 & 43.49 \\
CL-MoE & \faQuestionCircle / \faImage & 5000 & \textbf{46.50} & \textbf{37.18} & \textbf{75.22} & \textbf{71.39} & \textbf{40.90} & \textbf{69.54} & \textbf{43.66} & \textbf{52.68} & \textbf{55.55} & \textbf{20.74} & \textbf{52.96} \\
\midrule
\rowcolor{lightgreen}
\textbf{QUAD}    & \faQuestionCircle            & 5000 & \underline{35.87} & \underline{33.17} & \underline{66.93} & \underline{62.15} & \underline{34.09} & \underline{61.28} & \underline{35.03} & \underline{38.87} & \underline{48.66} & \underline{25.53} & \underline{44.16} \\
\bottomrule
\end{tabular}%
}
\vspace{-3mm}
\caption{\textbf{LLaVA-7B Performance}. Evaluation using the pretrained LLaVA-7B model.}
\label{tab:llava_performance}
\end{table}

\section{Out-of-Answer-Set Problem Evaluation}
\label{sec:out_of_answer_set}

To empirically analyze the \textit{out-of-answer-set problem}, we designed a controlled continual learning experiment within the VQACL setting. Our objective was to demonstrate how sequential fine-tuning without appropriate regularization leads to catastrophic forgetting, causing the model to misclassify previous-task questions by selecting answers from the current task’s answer space. This phenomenon, which we note is related to \textit{class recency bias} in Class-Incremental Learning (CIL)~\citep{rypesc2025task, mai2021supervised}, arises when the model disproportionately favors responses from the most recently learned task, even when answering questions about past tasks.

To assess this, we structured the training process into three sequential tasks: counting, action recognition, and color identification from VQAv2 dataset. Each task contained a fixed set of possible answers:
\begin{itemize}
    \item \textbf{Counting Task}: The model learned to predict numerical answers from the set \{One, Two, Three\}.
    \item \textbf{Action Recognition Task}: The model answered binary yes/no questions from the set \{Yes, No\}.
    \item \textbf{Color Identification Task}: The model identified object colors from the set \{Red, Blue, Green\}.
\end{itemize}

At each stage, the model was trained on the current task while being evaluated on all previous tasks to measure forgetting-induced answer space drift. For evaluation, we tested the model on 10 questions per task and verified whether its predicted answers belonged to the corresponding \textit{expected answer set} of the task. A misclassification was recorded if a model produced an answer outside the defined set, indicating that it had lost the ability to correctly respond using prior knowledge.

We compared two settings: (1) Sequential Fine-tuning (No Replay), where the model was updated on each new task without access to previous data, and (2) QUAD (Ours), which incorporated question-only replay to retain past knowledge with attention distillation. 

To quantify the severity of the \textit{out-of-answer-set problem}, we analyzed the \textit{prediction distribution shift} across tasks using confusion matrices. Specifically, we examined whether the model, when tested on past-task questions, incorrectly answered using responses restricted to the most recent task. For instance, a model fine-tuned on the \textit{color task} but evaluated on \textit{counting questions} was expected to misclassify numerical questions as colors (e.g., responding \textit{``Red''} instead of \textit{``Two''}). Similarly, after training on \textit{action} task, past counting questions were likely to be misclassified as \textit{``Yes''} or \textit{``No''}.

The results, visualized in Fig.\ref{fig.out_of_answer_set_problem}, revealed that sequential fine-tuning caused a stark shift in the prediction distribution, with nearly all responses aligning with the most recent task’s answer set. In contrast, \qstmethodshort{} mitigated this effect by preserving prior-task responses, demonstrating the effectiveness of question-only replay in preventing catastrophic forgetting without requiring image exemplars.

\section{Detailed Description of the VQACL Setting}
\label{sec:details_setting}

This section provides a detailed overview of the Visual Question Answering Continual Learning (VQACL) setting, as introduced by \cite{zhang2023vqacl}. The VQACL setting is designed to test a model's ability to generalise and retain knowledge across a sequence of tasks involving both visual and linguistic modalities, with a particular focus on compositional generalisation and knowledge retention.

The VQACL setting is organised into a two-level hierarchy of tasks that challenge both the visual and linguistic capabilities of the model.

\begin{table*}[ht]
\centering
\caption{Linguistic-driven task statistics of VQA v2 in the VQACL setting. Stan. Test denotes the standard test set.}
\label{tab:vqav2_stats}
\renewcommand{\arraystretch}{1.2}
\resizebox{\textwidth}{!}{%
\begin{tabular}{l|r|r|r|l}
\toprule
\textbf{Task}         & \textbf{Train} & \textbf{Val} & \textbf{Stan. Test} & \textbf{Examples} \\ \midrule
\textbf{Recognition}  & 131,478        & 5,579        & 5,628               & What is on the floor? What does the sign say? \\
\textbf{Location}     & 12,580         & 611          & 611                 & Where is the giraffe? Where are the people standing? \\
\textbf{Judge}        & 160,179        & 7,126        & 7,194               & Is the baby playing ball? Are the windows big? \\
\textbf{Commonsense}  & 25,211         & 1,114        & 1,100               & Do the elephants have tusks? Do the dogs know how to swim? \\
\textbf{Count}        & 62,156         & 2,651        & 2,658               & How many beds? How many seats are there? \\
\textbf{Action}       & 33,633         & 1,498        & 1,373               & Are they drinking wine? Is the person flying? \\
\textbf{Color}        & 50,872         & 2,322        & 2,192               & What color is the bedspread? What color are the gym shoes? \\
\textbf{Type}         & 23,932         & 1,119        & 1,089               & What type of building is this? What type of animal is shown? \\
\textbf{Subcategory}  & 31,594         & 1,477        & 1,416               & What brand is the umbrella? What brand are his shoes? \\
\textbf{Causal}       & 5,868          & 231          & 200                 & Why does he have glasses on? Why is the dog jumping? \\ \bottomrule
\end{tabular}
}
\end{table*}

\begin{table*}[ht]
\centering
\caption{Linguistic-driven task statistics of NExT-QA in the VQACL setting. Stan. Test denotes the standard test set. CW: CausalWhy; TN: TemporalNext; TC: TemporalCurrent; DL: DescriptiveLocation; DB: DescriptiveBinary; DC: DescriptiveCount; DO: DescriptiveOther; CH: CausalHow.}
\label{tab:nextqa_stats}
\renewcommand{\arraystretch}{1.2}  
\resizebox{\textwidth}{!}{%
\begin{tabular}{l|r|r|r|l}
\toprule
\textbf{Task}     & \textbf{Train} & \textbf{Val} & \textbf{Stan. Test} & \textbf{Examples} \\ \midrule
\textbf{CW}       & 13,552         & 1,928        & 3,333               & Why is the lady sitting down? Why is the baby's hair wet? \\
\textbf{TN}       & 5,685          & 895          & 1,399               & What does the baby do after picking up the toy? What did the lady do after adjusting the shirt? \\
\textbf{TC}       & 4,797          & 663          & 1,165               & What event is happening? What sport is the man doing? \\
\textbf{DL}       & 1,942          & 295          & 482                 & Where are the two people dancing? Where is this video taken? \\
\textbf{DB}       & 2,928          & 277          & 495                 & Is the baby able to walk? Does the girl cry? \\
\textbf{DC}       & 1,378          & 192          & 365                 & How many babies are there? How many dogs are there? \\
\textbf{DO}       & 2,549          & 356          & 672                 & What season is this? What does the man use to stir the food in the pan? \\
\textbf{CH}       & 4,400          & 683          & 1,174               & How did the singer project her voice? How did the boy in the box move forward? \\ \bottomrule
\end{tabular}
}
\end{table*}

\begin{table*}[ht]
\centering
\caption{Detailed information about the five object groups in VQA v2.}
\label{tab:vqa_groups}
\renewcommand{\arraystretch}{1.2}  
\resizebox{\textwidth}{!}{%
\begin{tabular}{l|l}
\toprule
\textbf{Task} & \textbf{Objects} \\ \midrule
\textbf{Group 1} & hot dog, fork, orange, snowboard, potted plant, person, toilet, laptop, surfboard, bench, bus, dog, knife, pizza, handbag, bicycle \\ \hline
\textbf{Group 2} & horse, cell phone, elephant, boat, zebra, apple, stop sign, microwave, spoon, cup, skateboard, tie, umbrella, sandwich, bear \\ \hline
\textbf{Group 3} & donut, truck, frisbee, giraffe, dining table, motorcycle, parking meter, car, oven, airplane, bed, sheep, baseball bat \\ \hline
\textbf{Group 4} & skis, baseball glove, tennis racket, tv, traffic light, kite, cake, keyboard, bottle, remote, bird, carrot \\ \hline
\textbf{Group 5} & suitcase, couch, broccoli, cow, fire hydrant, chair, mouse, cat, banana, wine glass, backpack, bowl, sports ball, train \\ \bottomrule
\end{tabular}%
}
\end{table*}

\begin{table*}[ht]
\centering
\caption{Detailed information about the five object groups in NExT-QA.}
\label{tab:nextqa_groups}
\renewcommand{\arraystretch}{1.2}  
\resizebox{\textwidth}{!}{%
\begin{tabular}{l|l}
\toprule
\textbf{Task} & \textbf{Objects} \\ \midrule
\textbf{Group 1} & bicycle, camel, bat, microwave, snake, sofa, traffic light, hamster/rat, chicken, oven, stop sign, vegetables, skateboard, bird, toilet, racket \\ \hline
\textbf{Group 2} & crab, camera, lion, ball/sports ball, crocodile, screen/monitor, baby walker, cat, squirrel, frisbee, cattle/cow, sheep/goat, adult, scooter, electric fan, stool \\ \hline
\textbf{Group 3} & piano, watercraft, kangaroo, train, fruits, pig, suitcase, bear, tiger, bench, elephant, motorcycle, horse, snowboard, surfboard, handbag \\ \hline
\textbf{Group 4} & ski, stingray, antelope, toy, child, duck, guitar, dish, fish, cake, turtle, leopard, laptop, panda, table, cup \\ \hline
\textbf{Group 5} & penguin, faucet, car, bottle, bus/truck, aircraft, baby, bread, baby seat, cellphone, sink, rabbit, backpack, chair, dog, refrigerator \\ \bottomrule
\end{tabular}%
}
\end{table*}

\begin{itemize}
    \item \textbf{Linguistically-Driven Tasks. }At the higher level, the VQACL setting comprises a series of linguistically-driven tasks, denoted as ${\mathcal{T}^1, \dots, \mathcal{T}^T}$, where $T$ represents the total number of tasks. Each task focuses on a specific reasoning skill, such as counting or color identification, and is characterized by a particular type of question. For example, a task focused on counting might involve questions beginning with "\textit{How many}" or "\textit{What number}". In our experiments, the VQAv2 dataset consists of $T\!=\!10$ such tasks, while the NExT-QA dataset includes $T\!=\!8$ tasks.

    \item \textbf{Visually-Driven Subtasks. }Nested within each linguistically-driven task are a series of visually-driven subtasks ${\mathcal{S}_1^t, \dots, \mathcal{S}_K^t}$. Each visually-driven subtask is associated with a specific object group $G_k$, formed by partitioning the total set of object classes ${\{c_i}\}_{i=1}^{C}$ into $K$ groups. These groups are then randomly assigned to different subtasks within each linguistic-driven task. In our implementation, both the VQAv2 and NExT-QA datasets are divided into $K\!=\!5$ visual subtasks, covering a total of $C\!=\!80$ object classes, following the categorization used in the COCO dataset \citep{cocodataset}.

    \item \textbf{Novel Composition Testing. }The VQACL setting also includes a novel composition testing process, designed to evaluate the model's compositional generalization abilities—its capacity to apply learned concepts to new combinations of objects and questions.
\end{itemize}

\noindent\textbf{Training and Testing Procedure.} During training, the model is exposed to a subset of the visual-driven subtasks within each linguistically-driven task. Specifically, one visual-driven subtask $\mathcal{S}_k^v$ is randomly excluded from the training phase for each linguistic-driven task. This excluded subtask is reserved for testing and serves as a novel composition, where the model must answer questions about unseen combinations of objects and reasoning skills.

\noindent\textbf{Cross-Validation and Fair Testing. }To ensure a fair evaluation of the model's generalization capabilities, the VQACL setting employs a $K$-fold object-independent cross-validation process. This involves repeating the training and testing procedure $K$ times, each time excluding a different visual-driven subtask. This ensures that the model encounters all object classes across different folds, thereby providing a comprehensive assessment of its ability to generalize to new combinations of objects and tasks.

\noindent\textbf{Continual Learning Challenges.} The VQACL setting presents a significant challenge for continual learning models, requiring them to balance the retention of knowledge from previously learnt tasks (stability) with the ability to adapt to new, continually arriving tasks (plasticity). By structuring tasks to involve both new and previously encountered concepts, the VQACL setting effectively tests the model's ability to minimize catastrophic forgetting while enabling knowledge transfer across tasks.

\section{Details of Evaluation Datasets}
\label{sec:datasets_order}
In this section, we provide a detailed overview of the two datasets used in our evaluation: VQA v2 and NExT-QA. Each dataset has been carefully structured into different tasks, which are used to evaluate the performance of our continual learning models.

We summarize the statistics of each dataset, focusing on both linguistic and object-related tasks. Tables \ref{tab:vqav2_stats} and \ref{tab:nextqa_stats} (previously described) present the linguistic-driven task breakdown, including categories such as \textit{Recognition}, \textit{Commonsense}, \textit{Count}, and others.

Additionally, we grouped the objects in each dataset into five distinct object groups to facilitate better understanding and comparison of the models' object recognition capabilities. Tables \ref{tab:vqa_groups} and \ref{tab:nextqa_groups} offer a detailed breakdown of the objects associated with each group in VQA v2 and NExT-QA, respectively. This categorization will aid in analyzing how the models perform across different object categories.

These two datasets, each structured uniquely in terms of linguistic tasks and object types, allow us to rigorously assess the models in varied real-world scenarios. Together, these benchmarks enable a comprehensive evaluation of the continual learning approaches proposed in this work.

\section{Extended Analysis of Plasticity/Stability Trade-Off}
\label{extended_analysis}
\begin{figure*}
    \centering
    \includegraphics[width=\textwidth]{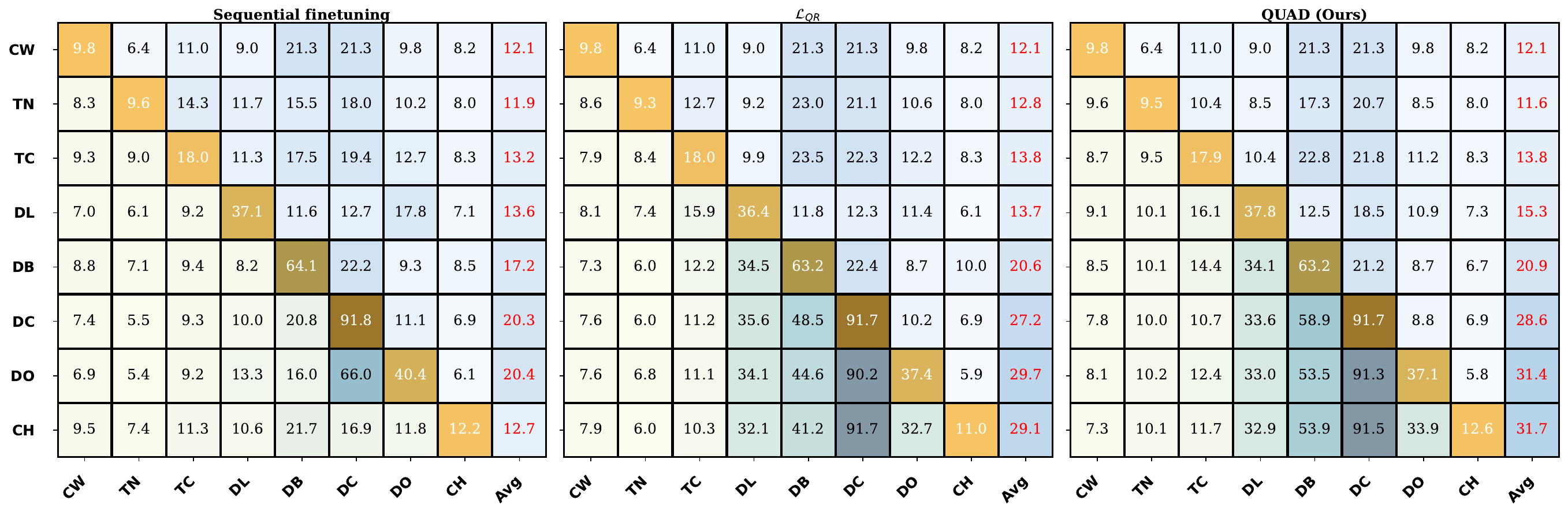}
    \caption{Comparison of feature distillation methods on \textbf{NExT-QA}. Each matrix shows the performance of a model trained on tasks (rows) and evaluated on tasks (columns). The diagonal (highlighted in {\color{orange}{orange}}) represents in-domain performance, while off-diagonal elements show cross-domain generalization. Higher values (darker colors) indicate better performance.}
    \label{fig:cross_task_generalization_nextqa}
\end{figure*}

Fig.\ref{fig:cross_task_generalization_nextqa} compares the impact of three continual learning strategies on performance across tasks in the NExT-QA dataset. The sequential finetuning baseline (left) demonstrates severe forgetting, with consistently low off-diagonal values. Specifically, tasks like temporal reasoning (TN and TC) exhibit the worst performance, as these tasks require advanced reasoning over time sequences, which is inherently challenging for the model.

Introducing pseudo-label distillation through $\mathcal{L}_{\text{PL}}$ (center) mitigates the issue of forgetting by enforcing output consistency with the previous model. This results in improved cross-domain retention, particularly in easier tasks like `DB' and `DL'. However, its performance on complex tasks such as "DO" (Descriptive Others) and `CH' (Causal How) remains suboptimal, as these tasks require the model to maintain intricate visual-linguistic relationships, which $\mathcal{L}_{\text{PL}}$ alone struggles to address.

Our method, \qstmethodshort{} (right), achieves the highest overall performance by combining pseudo-labeling with attention consistency distillation. This dual mechanism effectively balances stability and plasticity, as evidenced by the consistently high diagonal values and substantial improvements in off-diagonal cross-domain generalization. Notably, \qstmethodshort{} performs significantly better on retraining prior knowledge (row 6, 7). The results underscore the strength of \qstmethodshort{} in preserving visual-linguistic associations and mitigating the \textit{out-of-answer-set problem} across tasks in NExT-QA.

\section{Continual Learning Methods}
\label{explain_methods}
We assess and benchmark five prominent continual learning methods, encompassing two regularization techniques (EWC \citep{kirkpatrick2017overcoming}, MAS \citep{mas2018}) and three rehearsal-based methods (ER \citep{chaudhry2019tinyepisodicmemoriescontinual}, DER \citep{Chaudhry2019er}, VS \citep{Wan_2022_CVPR}, and VQACL\citep{zhang2023vqacl}). To ensure a consistent evaluation, all methods are implemented using their official codebases and integrated into the same transformer backbone as described in Section 5.1. 

\textbf{EWC} \citep{kirkpatrick2017overcoming} is a regularization method designed to preserve knowledge of prior tasks by selectively reducing updates on critical parameters. This is achieved by leveraging the Fisher Information Matrix, which quantifies the importance of parameters and incorporates an auxiliary L2 loss between significant parameters from old and new tasks.

\textbf{MAS} \citep{mas2018} similarly applies regularization, aiming to prevent significant changes to parameters vital for previous tasks by introducing an L2 loss. In contrast to EWC, MAS measures the sensitivity of the output with respect to parameter perturbations to estimate parameter importance.

\textbf{ER} \citep{chaudhry2019tinyepisodicmemoriescontinual} is a rehearsal method that utilizes a fixed-size memory buffer, where visited examples are stored and randomly sampled for retraining. In line with our approach, the memory size for ER is fixed at 5,000 for VQA v2 and 500 for NExT-QA. Given its simplicity and effectiveness, ER serves as the baseline for our proposed method.

\textbf{DER} \citep{Chaudhry2019er} is another rehearsal technique that employs reservoir sampling to manage memory, ensuring every visited sample has an equal chance of being stored. DER also incorporates a dark knowledge distillation strategy, which aims to align the network’s outputs with logits recorded during training, thus encouraging consistency in responses to prior examples. In our experiments, DER also utilizes memory sizes of 5,000 for VQA v2 and 500 for NExT-QA.

\textbf{VS} \citep{Wan_2022_CVPR} is a rehearsal-based method that emphasizes feature consistency between current and past data. To address forgetting, VS introduces two losses: a neighbor-session model coherence loss and an inter-session data coherence loss. For more details, we refer readers to \citet{Wan_2022_CVPR}. The memory size for VS is similarly set to 5,000 for VQA v2 and 500 for NExT-QA.

\textbf{VQACL} \citep{zhang2023vqacl} represents a rehearsal-based approach, incorporating a prototype module to learn both task-specific and invariant features, facilitating robust and generalizable representations for VQA tasks.

\end{document}